\documentclass{article}

\PassOptionsToPackage{numbers,sort&compress}{natbib}

    \usepackage[preprint]{neurips_2025}

\usepackage[utf8]{inputenc} 
\usepackage[T1]{fontenc}    
\usepackage{hyperref}       
\usepackage{url}            
\usepackage{booktabs}       
\usepackage{amsfonts}       
\usepackage{nicefrac}       
\usepackage{microtype}      
\usepackage{xcolor}         
\usepackage{amsmath}
\usepackage{ulem}
\usepackage{multirow} 
\usepackage{graphicx}
\usepackage{makecell}
\usepackage{hyperref}
\usepackage[capitalize]{cleveref}
\usepackage{subcaption}
\usepackage{color}
\usepackage{enumitem}
\usepackage{makecell}  
\usepackage{array}     
\usepackage{amssymb}
\usepackage{ulem}

\title{GRE Suite: Geo-localization Inference via Fine-Tuned Vision-Language Models and Enhanced Reasoning Chains}

\author{
    Chun Wang$^{1,2}$, Xiaojun Ye$^1$, Xiaoran Pan$^1$, \textbf{Zihao Pan}$^3$, \textbf{Haofan Wang}$^4$, \textbf{Yiren Song}$^{2,5}$\thanks{Corresponding author : songyiren725@gmail.com. }\\
    $^1$Zhejiang University, $^2$Creatly.ai, $^3$Sun Yat-sen University, $^4$LibLib.ai, $^5$NUS  \\
    \texttt{\{chunwang0326,songyiren725\}@gmail.com} \\
}

\begin{document}

\maketitle

\begin{abstract}
    Recent advances in Visual Language Models (VLMs) have demonstrated exceptional performance in visual reasoning tasks. However, geo-localization presents unique challenges, requiring the extraction of multigranular visual cues from images and their integration with external world knowledge for systematic reasoning. Current approaches to geo-localization tasks often lack robust reasoning mechanisms and explainability, limiting their effectiveness. To address these limitations, we propose the \textbf{G}eo \textbf{R}eason \textbf{E}nhancement (\textbf{GRE}) Suite, a novel framework that augments VLMs with structured reasoning chains for accurate and interpretable location inference. The \textbf{GRE} Suite is systematically developed across three key dimensions: dataset, model, and benchmark. First, we introduce \textbf{GRE30K}, a high-quality geo-localization reasoning dataset designed to facilitate fine-grained visual and contextual analysis. Next, we present the \textbf{GRE} model, which employs a multi-stage reasoning strategy to progressively infer scene attributes, local details, and semantic features, thereby narrowing down potential geographic regions with enhanced precision. Finally, we construct the \textbf{G}eo \textbf{R}eason \textbf{E}valuation Benchmark (\textbf{GREval-Bench}), a comprehensive evaluation framework that assesses VLMs across diverse urban, natural, and landmark scenes to measure both coarse-grained (e.g., country, continent) and fine-grained (e.g., city, street) localization performance. Experimental results demonstrate that \textbf{GRE} significantly outperforms existing methods across all granularities of geo-localization tasks, underscoring the efficacy of reasoning-augmented VLMs in complex geographic inference. Code and data will be released at \url{https://github.com/Thorin215/GRE}.
\end{abstract}

\section{Introduction}
\label{intro}
\vspace{-3mm}
Worldwide image geo-localization~\cite{vo2017revisiting,pramanick2022world} aims to predict the geographical coordinates of the shooting location based on any given photo taken anywhere on Earth. 
Unlike geo-localization within specific regions~\cite{noh2017large,tan2021instance,lee2022correlation}, global geo-localization, unrestricted to any specific region but covering the entire Earth, greatly unleashes the potential of geo-localization, which has significant applications across multiple domains, such as autonomous driving system positioning, social media image geo-tagging, and cultural heritage preservation.
However, precise global-scale image geo-localization still faces substantial technical challenges due to the vast diversity of global geographical environments, visual ambiguity between similar locations, and the variability of shooting conditions including weather patterns, seasonal changes, and lighting conditions.

\begin{figure}[!t]
  \centering
  \includegraphics[width=\linewidth]{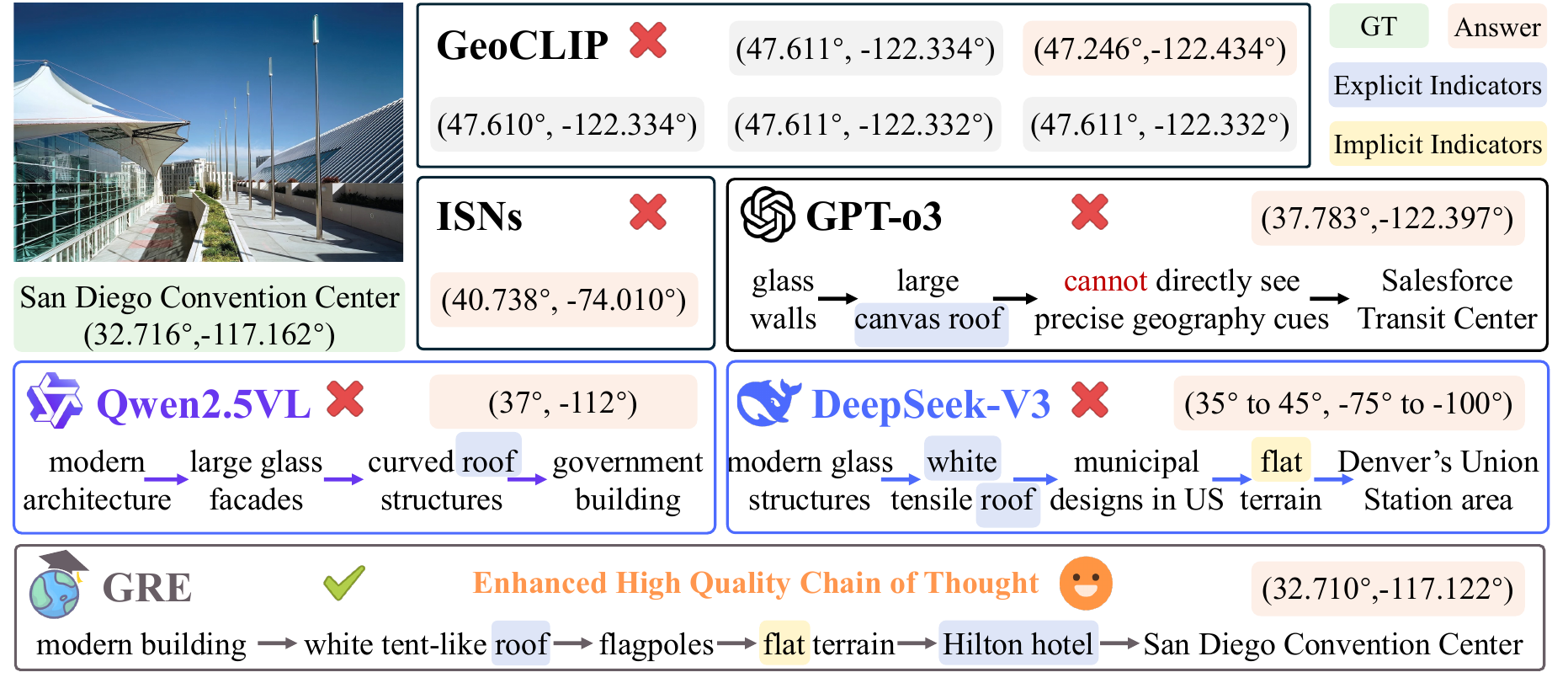}
  \caption{Performance comparison of our reasoning-based GRE versus traditional alignment-based approaches and MLLM baselines on image geo-localization.}
  \label{fig:qualitative}
\end{figure}

Geo-localization requires predicting the geographic coordinates of a photograph solely from the ground-view image. Extracting general geographical visual semantics is insufficient for the task, as two distant locations could potentially share similar image-level features. Instead, models need to \textbf{identify} and \textbf{reason} with geographically relevant visual elements from complex visual information. As illustrated in ~\cref{fig:qualitative}, when inferring the target location - San Diego Convention Center, the model is expected to jointly leverage explicit indicators such as the ``white sail'' roof design and implicit indicators such as flat terrain. However, existing approaches~\cite{vivanco2023geoclip,jia2024g3} rely on data-driven cross-modal alignment strategies, which establish correspondences through large-scale annotated image-GPS pairs while neglecting the inherent logical relationships among fine-grained geographical indicators within images. In addition, models need to predict geographic coordinates for images captured at any location in the world. However, existing methods based on closed-domain assumptions either maintain a candidate database of GPS coordinates~\cite{zhou2024img2loc,jia2024g3} or images~\cite{workman2015wide,liu2019lending,yang2021cross,zhu2022transgeo,tian2017cross}, or divide the entire geographical space into fixed grids for classification purposes~\cite{pramanick2022world,muller2018geolocation,clark2023we,izbicki2020exploiting,vivanco2023geoclip}, compromising the continuity of coordinate prediction. Thus, it is essential for image geo-localization models to possess the ability to predict \textbf{open-ended coordinates} without relying on candidate information, a feature that current methods inadequately address.


Recently, DeepSeek-R1~\cite{DeepSeek-R1} has successfully applied Reinforcement Learning (RL) to induce the self-emergence of complex cognitive reasoning ability in LLMs. Image geolocalization is inherently a multi-step cognitive process that requires progressive reasoning - from identifying visual cues in images, to inferring geographical correlations among these cues, and ultimately determining specific locations. This progressive reasoning process aligns naturally with the sequential decision-making characteristics of RL. Through RL, models can learn to formulate optimal reasoning strategies based on identified visual features, gradually narrowing down potential geographical regions, and ultimately arriving at accurate location predictions, rather than simply relying on pre-established image-GPS correspondences. Unfortunately, this direct RL training is challenged, as it struggles to effectively guide MLLMs generating complex CoT reasoning in absence of large-scale, high-quality multimodal data and prolonged training~\cite{huang2025vision}. 
What's more, fine-grained analysis of intermediate reasoning processes has proved beneficial for both evaluating and further improving models' reasoning capabilities~\cite{yang2025r1,jiang2025mme}. However, existing image geo-localization benchmarks~\cite{4587784,clark2023we} focus solely on terminal prediction accuracy while ignoring reasoning quality assessment.


To address the aforementioned challenges, we propose \textbf{G}eo \textbf{R}eason \textbf{E}nhancement (\textbf{GRE}), a novel reasoning solution that integrates cold-start supervised fine-tuning and two-stage reinforcement learning training for worldwide image geolocalization.
To facilitate the training process, we establish a geography reasoning dataset \textbf{GRE30k} by leveraging o3 to generate chain-of-thought demonstrations for geography seed questions. Our curated GRE30K consists of two sub-datasets: GRE30K-CoT, which contains format-standardized CoT content and answers refined through annotator filtering, and GRE30K-Judge, which comprises reasoning chain judgment tasks constructed through regular expression matching.
GRE30k-CoT serves as a cold start dataset to establish basic reasoning capabilities of the base model. Then, we need to apply two-stage Group Relative Policy Optimization (GRPO) ~\cite{DeepSeek-R1,deepseekmath} on a GRE30K-Judge and seed questions to enhance the model’s reasoning capability. 

Furthermore, to rigorously assess models' ability to leverage geographical visual cues for geo-localization and evaluate the quality of their reasoning chains, we develop a benchmark named \textbf{G}eo \textbf{R}eason \textbf{E}valuation Benchmark (\textbf{GREval-Bench}). Specifically, we design an automated pipeline to filter images containing geographical indicators and provide each image with a corpus of explicit and implicit geographical identifiers along with high-quality CoT annotations. 
We summarize the key contributions of our work as follows:


\begin{itemize}[leftmargin=*]
    \item We present \textbf{GRE}, a novel reasoning solution for the worldwide image geo-localization task. Our proposed methodology integrates cold-start initialization with a two-stage reinforcement learning training paradigm to effectively leverage geographical indicators within images and enable open-ended geolocalization. 
    
    \item We introduce \textbf{GER30K}, comprising a high-quality CoT dataset and a judgement task dataset. We anticipate the dataset will benefit more future work for location-aware visual reasoning.
    
    \item Furthermore, to comprehensively evaluate the image geo-localization capability of the models, we develop \textbf{GREval-Bench}, consisting of higher quality images, CoT quality assessments, and a corpus of geographic indicators.
\end{itemize}

\section{Related Work}
\vspace{-3mm}
\begin{figure}[!t]
  \centering
  \includegraphics[width=\linewidth]{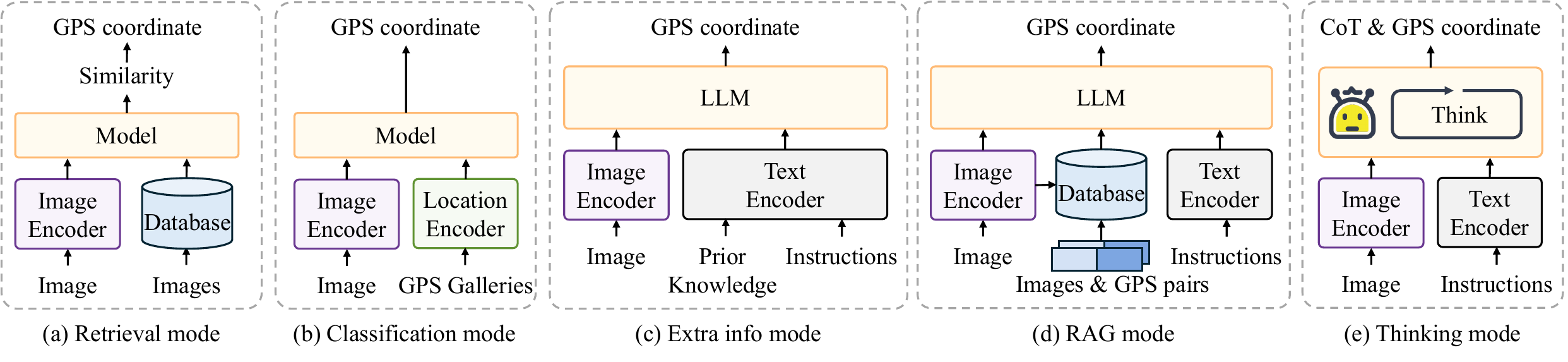}
  \caption{Summary of current image geo-localization model architectures.}
  \label{fig:summary}
\end{figure}

\textbf{Image Geo-localization.} Image Geo-localization is an important task in computer vision~\cite{zhu2023difftraj,zhu2023synmob,zhu2024controltraj}, spatial data mining~\cite{zhang2023promptst}, and GeoAI~\cite{zhao2022multi}. As shown in ~\cref{fig:summary}, previous work in image geo-localization can be divided into four main modes: classification mode, retrieval mode, prior knowledge mode and RAG mode. 
(1) Retrieval mode treat the image geo-localization task as a retrieval problem, typically maintaining a database of images~\cite{workman2015wide,liu2019lending,zhu2021vigor,yang2021cross,zhu2022transgeo,tian2017cross} or a gallery of GPS coordinates~\cite{vivanco2023geoclip}. They take the most similar images and GPS coordinates to the query image as the predicted values. However, maintaining a global-level image database or GPS gallery is infeasible. 
(2) Classification mode~\cite{seo2018cplanet,vo2017revisiting,muller2018geolocation,weyand2016planet,pramanick2022world,clark2023we} divide the entire earth into multiple grid cells and assign the center coordinates as predicted values. Models are then trained to classify the input image into the correct cell. However, if the actual location of the image is far from the center of the predicted cell, there can still be significant errors, even if the cell prediction is correct. 
(3) Prior Knowledge mode approaches~\cite{vivanco2023geoclip} incorporate higher-level geographical information, such as continental-scale priors, to enhance performance. Nevertheless, this approach essentially provides partial solutions, contradicting the fundamental purpose of the task.
(4) RAG mode~\cite{zhou2024img2loc,jia2024g3} leverage large language models by retrieving relevant image-GPS pairs as references to optimize predictions. While there are also some tries based on diffusion method like \textit{Around the World}~\cite{11094630}, with application of flow matching and diffusion~\cite{lipman2023flowmatchinggenerativemodeling, zhang2025isdrama, zhang2025easycontrol, zhang2024ssr, song2025layertracer, huang2025psdiffusionharmonizedmultilayerimage} . However, these approaches rely on establishing large-scale aligned databases. In contrast to existing global image geo-localization approaches, we propose a reasoning-based methodology that leverages both explicit and implicit geographical indicators within images to predict open-ended coordinate prediction. Recent advances in MLLMs have enabled novel approaches leveraging their reasoning capabilities for geographic inference. While some works ~\cite{dou2025gagainteractiveglobalgeolocation, seekworld2025, li2024georeasoner}employ explicit reasoning chains, they lack systematic evaluation of reasoning quality. Complementary work has developed datasets ~\cite{10657636, song2025geolocationrealhumangameplay} and reinforcement learning frameworks ~\cite{seekworld2025} to enhance human-like geospatial reasoning.

\textbf{Vision Language Models (VLMs).} Models in the vein of GPT-4o~\cite{2024gpt4o} achieve excellent visual understanding ability by integrating both visual and textual data. This integration enhances the models' ability to understand complex multi-modal inputs and enables more advanced AI systems~\cite{wang2024qwen2,li2024llavaov,zhang2024internlm,liu2024deepseekv3} capable of processing and responding to both images and text. Generally, the training of LVLMs involves two steps: (a) pre-training and (b) post-training which contains supervised fine-tuning and reinforcement learning. Post-training is crucial in improving the model's response quality, instruction following, and reasoning abilities. While there has been significant research on using reinforcement learning to enhance LLMs during post-training~\cite{lm-human-preferences, Learning-to-summarize, Training-language-models, RL4LMs, zang2024contextual, Grounding-large-language-models, Aligning-LLMs-with-RLHF, ILQL, LMRL-Gym, ArCHer, ReAct}, the progress for LVLMs has been slower. In this paper, we propose GRE-RL, which used GRPO-based reinforcement algorithms and verifiable reward during the post-training phase to enhance the model’s visual perception and reasoning capabilities.

\textbf{Reinforcement Learning.} Recently, with the emergence of reasoning models like OpenAI's o1~\cite{OpenAI_O1} and Deepseek-R1~\cite{DeepSeek-R1}, the research focus in Large Language Models (LLMs) has increasingly shifted towards enhancing the models' reasoning capabilities through reinforcement learning (RL) techniques. Studies have explored improving LLMs' performance in reasoning tasks such as solving mathematical problems~\cite{deepseekmath,yang2024qwen2math,ying2024internlmmath,cai2024internlm2,luong2024reft} and coding~\cite{hui2024qwen2coder,jiao2024preferencecode,zhang2024o1,zhang2024codedpo}. A notable breakthrough in this area is Deepseek-R1-Zero~\cite{DeepSeek-R1}, which introduced a new approach to achieving robust reasoning capabilities using RL merely, eliminating the supervised fine-tuning (SFT) stage. However, current research on RL-based reasoning has largely been confined to the language domain, with limited exploration of its application in multi-modal settings. For LVLMs, RL has primarily been used for tasks like mitigating hallucinations and aligning models with human preference~\cite{hadpo,yu2024rlhfv,liu2024mia,zhou2024aligning, Ye2025M4BenchAB, zhang2024gtsinger, song2025makeanything, ye2025focusedad, ye2024mmad}. Interpretable visual reasoning, once a longstanding challenge~\cite{HE2021104194}, now benefits from RL-finetuned LVLMs acting as decision agents~\cite{RL4VLM}. Cutting-edge models like Kimi~\cite{kimiteam2025kimivltechnicalreport} demonstrate advanced capabilities, with research expanding beyond hallucination mitigation to core reasoning enhancement. However, there remains a significant gap in research focusing on enhancing reasoning and visual perception of Large Vision Language Models. To address this gap, our work uses a novel reinforcement fine-tuning strategy , applying verifiable rewards with GRPO-based~\cite{deepseekmath} RL to visual geo-localization tasks. Our approach aims to improve the performance of LVLMs in processing various geo-localization tasks, especially when the high-quality fine-tuning data is limited.

\section{Methodology}
\label{sec:method}
\vspace{-3mm}
\begin{figure}[!t]
  \centering
  \includegraphics[width=\linewidth]{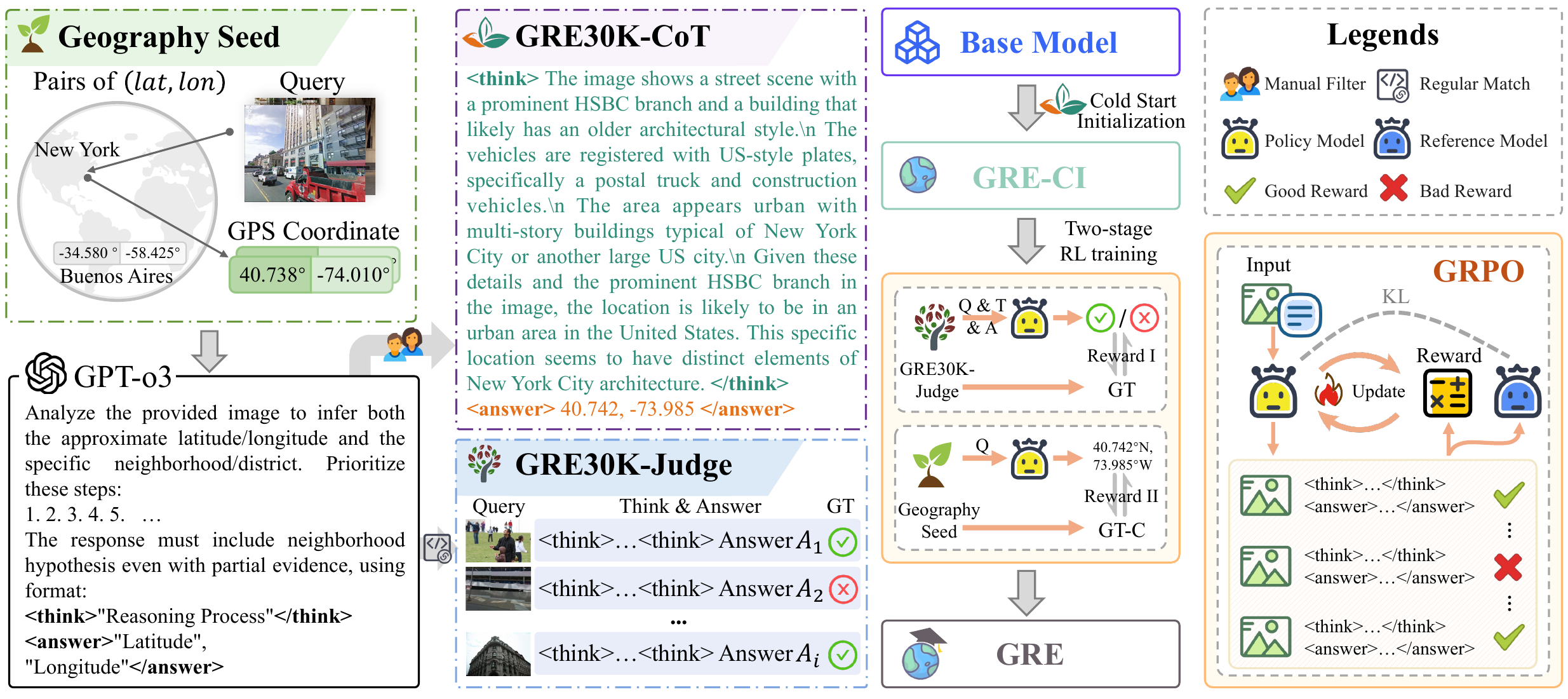}
  \caption{Overview of our GRE framework. The geographical reasoning pipeline begins with data preparation, incorporating automated CoT generation, regular expression matching, and manual filtering. Based on our constructed GRE30K dataset, we employ a post-training procedure that consists of supervised fine-tuning to learn reasoning patterns, followed by two-stage rule-based reinforcement learning to enhance image geo-localization reasoning capabilities.}
  \label{fig:architecture}
\end{figure}

~\cref{fig:architecture} illustrates the comprehensive reasoning pipeline of GRE.
This method begins with a cold-start using a high-quality geo-localization Chain-of-Thought dataset, which initially teaches the base model to reason step-by-step following human-like patterns. Subsequently, we apply a two-stage reinforcement learning training to the cold-start initialized model GRE-CI to guide it towards adopting the correct geographical reasoning process, thereby enhancing the geo-localization reasoning capability in the final model GRE.

In the following sections, we first describe our approach to create a high-quality geo-localization reasoning dataset GRE30K in ~\cref{sec:data}. Then we introduce our proposed Post-Training Strategy, comprising cold-start supervised fine-tuning (~\cref{sec:SFT}) and two-stage reinforcement learning training (~\cref{sec:RL}). Correspondingly, our GRPO-based training strategy and two-stage reward function design will be described in ~\cref{sec:reward}.


\subsection{GRE30K Construction}
\vspace{-2mm}
\label{sec:data}
In this section, we present GRE30K, a geo-localization reasoning dataset designed to enhance the visual reasoning capability of MLLMs. Specifically, GRE30K consists of GRE30K-CoT for cold-start Initialization and GRE30K-Judge for reinforcement learning. 
Examples of the generated data are provided in ~\cref{sec:gre30k_examples}. While GWS15k~\cite{clark2023we} reveals Im2GPS3k~\cite{4587784}'s non-uniform distribution (with landmark repetition risks), our geographic filtering ensures clean evaluation.

\paragraph{Reasoning Process Generation.}
We make full use of the publicly available dataset MP16-Pro~\cite{jia2024g3} with GPS coordinates. However, the source dataset only contains images, coordinates, and discrete geographical information including the corresponding county and state for each image, which are insufficient to train an MLLM. Our goal is to construct a CoT dataset that encompasses complex cognitive processes to facilitate our training strategy, enabling GRE to reason in a manner that closely resembles human cognitive patterns. Furthermore, GPT-o3 has demonstrated the capabilities in generating CoT reasoning that mirrors natural cognitive processes and has proven to have strong reasoning capability. Leveraging these insights, we employ GPT-o3 to generate image-CoT-coordinate triples through meticulously designed prompt templates. Please refer to ~\cref{sec:o3_prompt} for the detailed prompts for GPT-o3. 

\paragraph{GRE30K-CoT.} 
To address potential errors and mismatches in source CoT data, we combine automated filtering and manual verification to ensure the quality and reliability of the test data. Please refer to~\cref{sec:cot_details} for more details.
Finally, we collect 20k high-quality CoT samples.
By acquiring CoT data in this manner, which closely mimics human cognitive behavior, reasoning processes exhibit natural and logical thinking.

\paragraph{GRE30K-Judge.}
In addition to standardizing the model's reasoning process through high-quality CoT data, we develop GRE30K-Judge, a judgment task dataset. This dataset is created by comparing extracted predictions with ground truth using threshold $\theta$, labeling images as "Truth" or "False" accordingly. The resulting dataset is incorporated into reinforcement learning training, enabling the model to learn from both correct and incorrect reasoning patterns and thereby enhancing its geographical reasoning abilities. In total, we obtain 10k judgment samples.

\subsection{Post-Training Strategy}
\vspace{-2mm}
To enhance visual reasoning capabilities, we introduce a three-stage post-training strategy consisting of cold-start initialization and two-stage rule-based reinforcement learning (RL). SFT stabilizes the model’s reasoning process and standardizes its output format, while RL further improves generalization across various geo-localization tasks.

\subsubsection{Cold-start Initialization}
\label{sec:SFT}
Leveraging the GRE30K-CoT dataset, we conduct SFT on a pretrained MLLM as the base MLLM for cold-start initialization. The MLLM after cold start initialization is named as GRE-CI. At this stage, the base MLLM had learned the complex reasoning mode from o3~\cite{OpenAI_O3}. Through SFT with the GRE30K-CoT dataset, the model standardize output format and establish a systematic reasoning framework. This critical phase facilitates the model's acquisition of high-quality structured reasoning patterns, thereby constructing a solid foundation for subsequent RL procedures.

\subsubsection{Reinforcement Learning on the GRE-CI}
\label{sec:RL}
Building upon the SFT-trained model, we employ rule-based reinforcement learning (RL) to optimize structured reasoning and ensure output validity. Specifically, we define two kinds of reward rules inspired by R1 and update the model using Group Relative Policy Optimization (GRPO). The RL stage further encourages the model to generate reliable outputs and enhances its generalization capabilities in geographical reasoning tasks. Please refer to ~\cref{sec:rl_training} for more details about the two-stage RL training pipeline.

\paragraph{Rule-Based Rewards.}
We define two kinds of reward rules that evaluate the generated answers from two perspectives:
\begin{itemize}[leftmargin=*]
    \item \textbf{Accuracy Reward}: The accuracy reward rule evaluates the correctness of the final answer by extracting final answer via regular expressions and verifying them against the ground truth. For image geo-localization task, the final answer must be provided in a specified format to enable reliable rule-based verification. In \textbf{RL stage I}, given an input image along with its CoT and predicted answer, the model evaluates the correctness of both the reasoning process and the final answer. The model receives a reward score of $r_i = 1$ only if the generated final result aligns with the ground truth; otherwise, it receives a score of $r_i = 0$. In \textbf{RL stage II}, where the model directly predicts coordinates based on the input image, the reward is determined by the threshold metric $\theta$.
    \item \textbf{Format Reward}: In order to ensure the existence of the reasoning process, the format reward rule requires that the response must follow a strict format where the model’s reasoning is enclosed between \texttt{<think>} and \texttt{</think>}. A regular expression ensures the presence and correct ordering of these reasoning markers. What's more, \texttt{<answer>} and \texttt{</answer>} are used to ensure model have given a answer.
\end{itemize}

\subsection{Group Relative Policy Optimization}
\vspace{-2mm}
\label{sec:reward}
We employ GRPO to achieve balanced integration of consistent policy updates and robust reward signals in a controlled manner. For each token in the generated output, GRPO first compute the log probabilities under both the new policy ($\pi_{\theta}$) and a reference policy ($\pi_{\text{ref}}$). It then calculates the probability ratio and clips it to the range $[1-\epsilon, 1+\epsilon]$ to constrain policy updates and avoid divergence. The normalized reward (treated as an advantage estimate) is subsequently used in a PPO-style loss function, combining policy optimization with KL-divergence (weighted by $\beta$) regularization:
\begin{equation}
    \mathcal{L}_{\text{clip}} = -\mathbb{E}\Bigl[\min\Bigl(\text{ratio}_t \cdot \text{Adv}_t,\ \text{clipped\_ratio}_t \cdot \text{Adv}_t\Bigr)\Bigr].  
\end{equation}
\begin{equation}
\begin{split}
\mathcal{L}_{\text{GRPO}}(\theta) = -\mathbb{E}\Bigl[\, & \min\Bigl(\text{ratio}_t \cdot \text{Adv}_t,\ \text{clipped\_ratio}_t \cdot \text{Adv}_t\Bigr) \\
& - \beta \cdot \operatorname{KL}\Bigl(\pi_\theta(y\mid x),\ \pi_{\text{ref}}(y\mid x)\Bigr) \Bigr].  
\end{split}
\end{equation}
Here, \(\text{Adv}_t\) denotes the advantage function, capturing how much better (or worse) a particular action is compared to a baseline policy value.

Compared to other methods, the GRPO clipping mechanism prevents extreme policy shifts, while the KL regularization keeps the updated policy aligned with the baseline. This combination ensures that our model integrates rule-based rewards efficiently without compromising training stability.
Subsequently, we will introduce the reward function R adopted for second-stage(~\cref{eq:binary_reward}) and third-stage(~\cref{eq:geo_reward}). 

\begin{equation}
d = \text{geodesic}\big((\phi_{\text{pred}}, \lambda_{\text{pred}}), (\phi_{\text{true}}, \lambda_{\text{true}})\big) \\
\end{equation}
\begin{equation}
R_{\text{yes/no}}(y_{\text{pred}}, y_{\text{true}}) = 
\begin{cases}
1.0 & \text{if } \mathcal{E}(y_{\text{pred}}) = \mathcal{E}(y_{\text{true}}) \\
0.0 & \text{otherwise}
\end{cases}
\label{eq:binary_reward}
\end{equation}
\begin{equation}
R_{\text{geo}}(y_{\text{pred}}, y_{\text{true}}) = 
\begin{cases} 
\frac{2}{1 + \exp(d/\theta)} & \text{if } \mathcal{V}(y_{\text{pred}}, y_{\text{true}}) = \text{True} \\
0 & \text{otherwise}
\end{cases}
\label{eq:geo_reward}
\end{equation}

Here, \(\theta\) denotes the threshold, it is used as a factor to control the range of reward in this reward function ~\cref{eq:geo_reward}. $\mathcal{E}$ mean the boolean value of the prediction and $\mathcal{V}$ mean the values of  prediction and ground truth are valid.

\section{GREval-Bench}
\vspace{-3mm}

To comprehensively evaluate the image geo-localization capability of the models, we develop a geographical reasoning benchmark named \textbf{GREval-Bench}. Existing benchmarks~\cite{4587784,thomee2016yfcc100m} are directly constructed from geotagged Flickr images without appropriate filtering. Specifically, these benchmarks contain numerous images that lack geographical relevance cues, such as portraits and object-focused photographs. The inclusion of such geographically uninformative samples compromises the validity of evaluation results.
Moreover, these benchmarks primarily focus on final predictions while neglecting the evaluation of the entire CoT process. The CoT process reflects multiple aspects of geographical reasoning capabilities and serves as a critical medium for understanding models' reasoning patterns and limitations. 

To address these challenges, we propose an semi-automated pipeline for geo-localization image filtering and CoT annotation generation in our GREval-Bench. ~\cref{fig:pie} and ~\cref{tab:bench_statistic} provide data statistics, respectively. Please refer to ~\cref{sec:bench_details} for more details of the GREval-Bench construction and evaluation pipeline. GREval-Bench comprises 3K triplets, each containing: (1) geographical inference images filtered through our pipeline, (2) a corresponding corpus of geographical indicators categorized into explicit and implicit types, with detailed subcategories presented in ~\cref{sec:subcategories_details}, and (3) reference GPS coordinates and annotated key Chain-of-Thought steps, where step categories and partitioning follow ~\cite{jiang2025mme}. Through our construction pipeline, we have enhanced both the image quality and complexity of the benchmark by eliminating noisy images lacking geographical indicators while increasing the proportion of samples that require reasoning based on implicit indicators. This improvement facilitates a more accurate assessment of models' geo-localization capabilities.

\begin{minipage}{\textwidth}
  \begin{minipage}[b]{0.65\textwidth}
    \centering
    \centerline{\includegraphics[width=\textwidth]{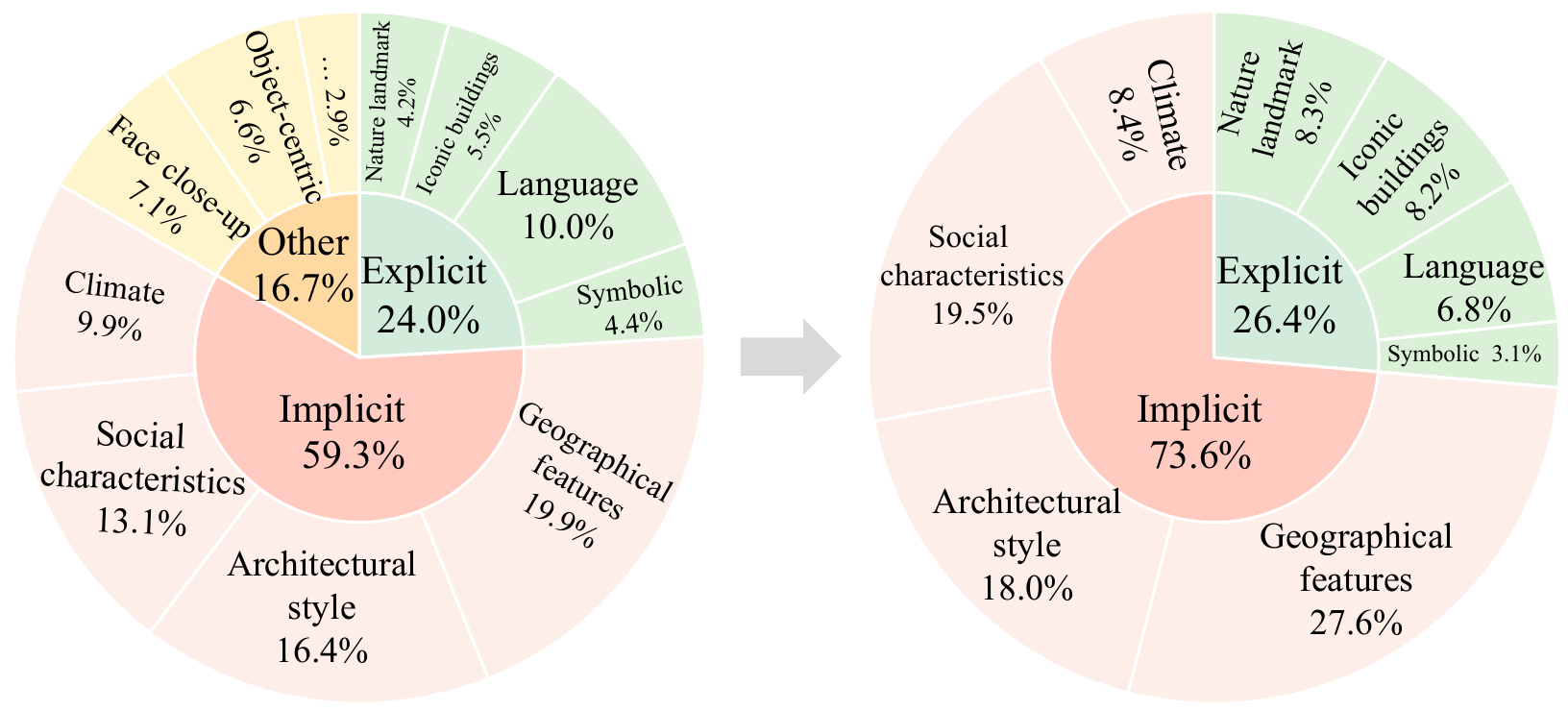}}
    \captionof{figure}{\small{Indicators distribution of GREval-Bench.}}
    \label{fig:pie}
  \end{minipage}
  \hfill
    \begin{minipage}[b]{0.34\textwidth}
    \centering
    \scalebox{0.8}{
    \tabcolsep=0.09cm
    \begin{tabular}{l c}  
    \hline
    \textbf{Statistic} & \textbf{Number} \\
    \hline  
    Outdoor & 2400 \\
    \quad - natural scene & 811 \\
    \quad - artificial landscape & 1138 \\
    \quad - agricultural scene & 58 \\
    \quad - industrial scene & 66 \\
    \quad - road traffic & 327 \\
    \hline
    Indoor & 600 \\
    \quad - commercial premises & 147 \\
    \quad - offices & 131 \\
    \quad - transportation place &  54\\
    \quad - cultural sites & 148 \\
    \quad - medical place & 4 \\
    \quad - entertainment venues & 116 \\
    
    \hline
    \end{tabular}
    }
    \captionof{table}{\small{Statistics of GREval-Bench.}}
    \label{tab:bench_statistic}
    \end{minipage}
  \end{minipage}

\begin{figure}[!t]
  \centering
  \includegraphics[width=\linewidth]{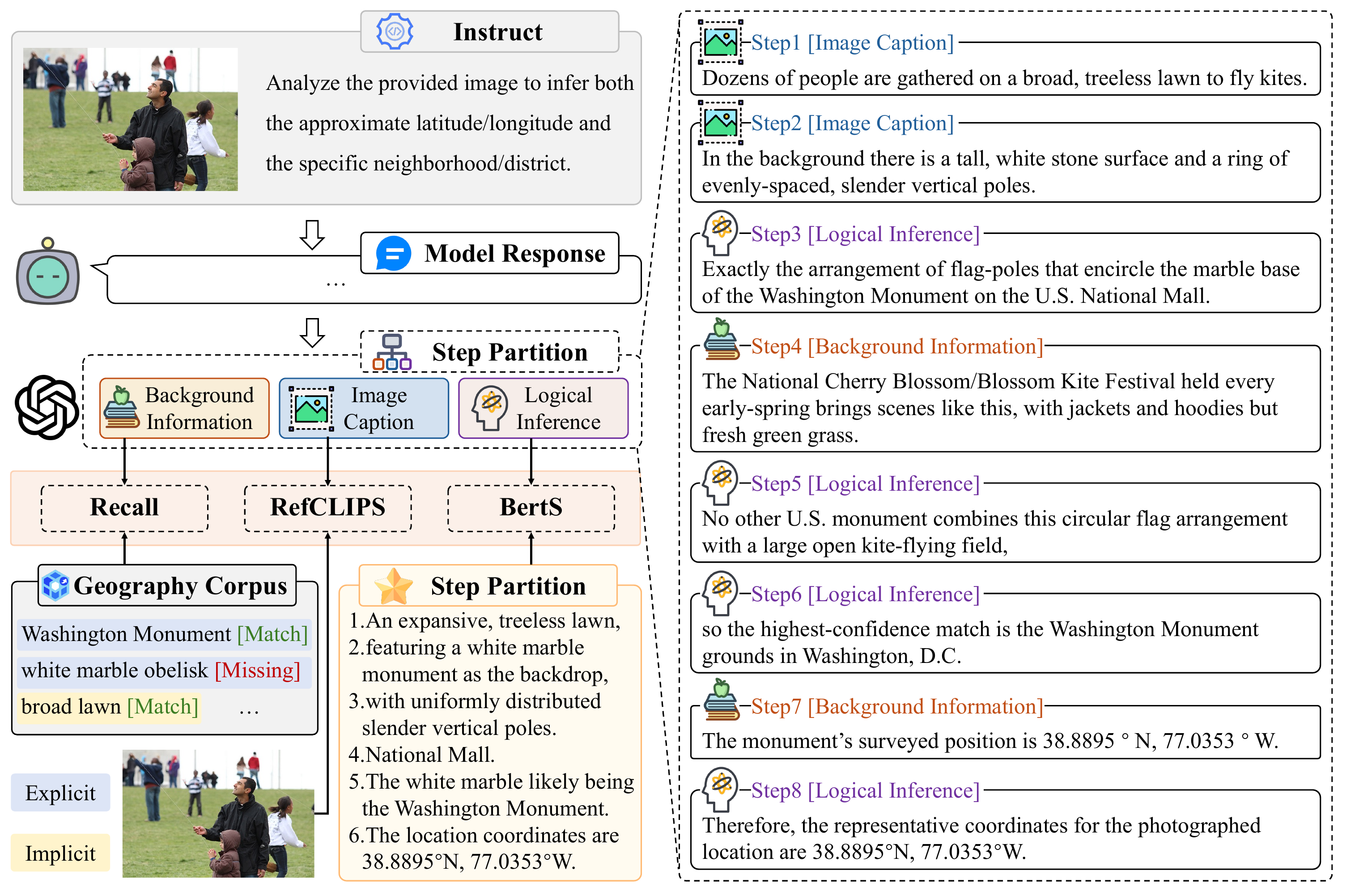}
  \caption{A detailed illustration of the evaluation pipeline.}
  \label{fig:bench_eval}
\end{figure}

As illustrated in ~\cref{fig:bench_eval}, we instruct GPT-4o~\cite{2024gpt4o} to categorize each reasoning step into three categories: background information, image caption, and logical inference. We calculate the recall between background information and the corresponding geography corpus. Then, we employ RefCLIPScore~\cite{hessel2021clipscore} to evaluate the semantic alignment between image captions and visual content, and utilize BertScore~\cite{zhang2019bertscore} to assess the similarity between predicted and ground-truth logical inference steps.
As these components are crucial for visual reasoning, we calculate CoT-quality by the follow equation (~\cref{Eq:cot-quality}). 
\begin{equation}
\text{CoT-quality} = \frac{\text{Recall} + \text{RefCLIPS} + \text{BertS}}{3}
\label{Eq:cot-quality}
\end{equation}

\section{Experiment}
\vspace{-3mm}
\label{sec:experiments}

\textbf{Datasets and Evaluation details}:
We randomly sample 5\% of MP-16~\cite{larson2017benchmarking}, a dataset containing 4.72 million geotagged images from Flickr~\footnote{https://www.flickr.com/}, as geography seed datasets to construct our GRE30K. This dataset is strategically utilized across our three-stage training process: GRE30K-CoT, comprising 20k high-quality Chain-of-Thought examples curated by geography experts and standardized in format, serves for cold-start initialization; GRE30K-Judge, consisting of 10k CoT judgment tasks, is employed for Stage I reinforcement learning training and the remaining 170k seed datasets are utilized for Stage II reinforcement learning training. We test our trained model on Im2GPS3k~\cite{4587784} and Google World Streets 15k (GWS15k)~\cite{clark2023we}. 
To ensure a fair comparison with existing methods in the evaluation of Im2GPS3k, both our proposed model and transformer-based models are trained using only 5\% of the MP-16 dataset.
Follow the protocol followed in previous works~\cite{jia2024g3,vivanco2023geoclip}, we report our results using a threshold metric. Given the predicted coordinates and the ground truths,
this metric quantifies the percentage of predictions where the distance to the ground truth falls within specified thresholds (1km, 25km, 200km, 750km, and 2500km).

\textbf{Implementation details}:
We adopt Qwen2.5-VL-7B as base model, the SFT experiments are conducted with a batch size of 128, a learning rate of 1e-5, and training over 1 epochs. Then, we perform RL on our dataset and experiment with training subsets of 10k for a single epoch each. All experiments are conducted with PyTorch and 8 NVIDIA H20(96G) GPUs.

\subsection{Comparison with State-of-the-art methods}
\vspace{-2mm}
We perform a comparative analysis of GRE against worldwide Geo-Localization benchmarks, Im2GPS3k and GWS15k. The results on Im2GPS3k~\cite{4587784} and GWS15k~\cite{clark2023we} are shown in ~\cref{tab:state-of-the-art}. In all metrics, our method surpasses the previous state-of-the-art (SOTA) model on Im2GPS3k, achieving improvements of +0.5\%, +4.2\%, +3.0\%, +1.7\% and +2.5\% in the 1km, 25km, 200km, 750km, and 2500km thresholds respectively. The results on additional geographical benchmarks are put in \cref{sec:additional_xperiment}, where we also observe a similar trend.

Moreover, our approach exhibits a large gain on the more challenging GWS15k dataset, surpassing the previous SOTA model with significant accuracy improvements of +0.2\%, +1.0\%, +2.0\%, and +4.2\% in the 1km, 25km, 200km and 2500km thresholds respectively. Our model achieves superior performance over previous state-of-the-art approaches while utilizing merely 5\% of the data, compared to their use of the complete MP-16 dataset. The GWS15k contains samples that are uniformly sampled across the Earth and are not biased towards any specific geographic location. Moreover, the images in this dataset have a large distribution shift compared to the training set, making the geo-localization task tough and challenging for brute-force alignment approaches. Our substantial improvement can be attributed to effective reasoning that leverages both explicit and implicit geographical indicators within images.

Geo-localization in Vision-Language Models (VLMs) indeed highlights their ability to integrate world knowledge for inference—an emergent capability developed during training. To provide a comprehensive comparison, we have benchmarked both LLaVA-1.5 ~\cite{liu2023visual} and Molmo-D-7B~\cite{molmo} on the Im2GPS3k dataset, which use open-source training data.

\begin{table}[htp]
\centering 
\vspace{-0.1in}
\caption{\small{We compare the performance of GRE with the state-of-the-art methods on (a) Im2GPS3k~\cite{4587784} and (b) GWS15k~\cite{clark2023we} datasets. Our method yields consistent gains across datasets and different distance thresholds. ${\dagger}$ denotes transformer-based models. The asterisk ($*$) signifies that for a direct comparison, GeoReasoner was prompted to output coordinates, which differs from its default city-name output format.}} 
    \begin{subtable}{.46\linewidth}
    \raggedright
    \caption{\centering{\small{Results on the Im2GPS3k~\cite{4587784} dataset}}}
    \scalebox{0.7}
    {
    \tabcolsep=0.09cm
    \begin{tabular}{c || c c c c c}
    \hline
    \multirow{2}{*}{\bf \centering Method} & {\bf \raggedleft Street} & {\bf \raggedleft City} &{ \bf \raggedleft Region} & { \bf \raggedleft Country} & { \bf \raggedleft Continent} \\ 
    
     & \bf $1$ km & \bf $25$ km & \bf $200$ km & \bf $750$ km & \bf $2500$ km \\
    
    \hline
    \hline

    \centering [L]kNN, $\sigma$ = $4$~\cite{vo2017revisiting} & 7.2 & 19.4 & 26.9 & 38.9 & 55.9\\
    
    \centering PlaNet~\cite{weyand2016planet} & 8.5 & 24.8 & 34.3 & 48.4 & 64.6\\
    
    \centering CPlaNet~\cite{seo2018cplanet} & 10.2 & 26.5 & 34.6 & 48.6 & 64.6\\
    
    \centering ISNs~\cite{muller2018geolocation} & 3.2 & 9.6 & 14.3 & 25.1 & 43.9\\  
    
    Translocator${^\dagger}$~\cite{pramanick2022world} & 7.6 & 20.3 & 27.1 & 40.7 & 63.3\\
    
    GeoDecoder${^\dagger}$~\cite{clark2023we} & 5.7 & 10.3 & 21.4 & 28.9 & 38.6\\
    
    GeoCLIP${^\dagger}$~\cite{vivanco2023geoclip} & 10.8 & 31.1 & 48.7 & 67.6 & 83.2\\

    GeoReasoner${^*}$~\cite{li2024georeasoner} & 0.2 & 1.6 & 2.1 & 3.9 & 6.8 \\
    
    GeoReasoner~\cite{li2024georeasoner} & 9.9 & 33.8 & 46.1 & 65.3 & 80.3 \\


    SeekWorld ~\cite{seekworld2025} & 4.3 & 29.8 & 44.9 & 59.1 & 67.3 \\

    Qwen2.5-VL-7B~\cite{bai2025qwen2} & 3.2 & 16.6 & 28.0 & 42.1 & 53.0 \\

    LLaVA-v1.5-7B~\cite{liu2023llava} & 1.7 & 7.5 & 11.3 & 20.8 & 44.6 \\

    Molmo-D-7B~\cite{molmo} & 2.1 & 9.8 & 19.6 & 36.3 & 55.7 \\

    \bf Ours & \bf 11.3 & \bf 35.3 & \bf 51.7 & \bf 69.3 & \bf 85.7\\
    
    \hline
    \end{tabular}
    }
    \end{subtable}
\hskip0.8cm
    \begin{subtable}{.46\linewidth}
    \raggedleft
    \caption{\centering{\small{Results on the recent GWS15k~\cite{clark2023we} dataset}}}
    \scalebox{0.7}
    {
    \tabcolsep=0.09cm
    \begin{tabular}{c || c c c c c}
    \hline
    \multirow{2}{*}{\bf \centering Method} & {\bf \raggedleft Street} & {\bf \raggedleft City} &{ \bf \raggedleft Region} & { \bf \raggedleft Country} & { \bf \raggedleft Continent} \\ 
    
     & \bf $1$ km & \bf $25$ km & \bf $200$ km & \bf $750$ km & \bf $2500$ km \\
    
    \hline
    \hline
    \multirow{4}{0.1cm}
    
    \centering ISNs~\cite{muller2018geolocation} & 0.05 & 0.6 & 4.2 & 15.5 & 38.5 \\  
    
    Translocator${^\dagger}$~\cite{pramanick2022world} & 0.5 & 1.1 & 8.0 & 25.5 & 48.3 \\
    
    GeoDecoder${^\dagger}$~\cite{clark2023we} & 0.7 & 1.5 & 8.7 & 26.9 & 50.5\\
    
    GeoCLIP${^\dagger}$~\cite{vivanco2023geoclip} & 0.6 & 3.1 & 16.9 & 45.7 & 74.1\\

    GeoReasoner${^*}$~\cite{li2024georeasoner} & 0.01 & 0.01 & 2.3 & 10.9 & 18.0 \\

    GeoReasoner~\cite{li2024georeasoner} & - & 0.9 & - & \underline{65.4} & - \\

    SeekWorld ~\cite{seekworld2025} & 0.2 & 1.9 & 9.5 & 34.1 & 67.3 \\
    
    \bf Ours & \bf 0.9 & \bf 4.1 & \bf 18.9 & 54.8 & \bf 78.3\\
    
    \hline
    \end{tabular}
    }
    \end{subtable}
\label{tab:state-of-the-art}

\end{table}
\subsection{Performance on GREval-Bench}
\vspace{-2mm}
We compare our approach on GREval-Bench with the previous generalist models,
including InternVL2.5 series~\cite{chen2024expanding}, InternVL3 series~\cite{zhu2025internvl3}, Qwen2.5-VL series~\cite{bai2025qwen2}. We conduct comprehensive evaluations of models, analyzing the above metric across different distance thresholds and scenarios, while also assessing the quality of its reasoning chains. ~\cref{tab:bench} presents the comparison results. Our approach achieves the leading average performance in various evaluation metrics while demonstrating more coherent reasoning processes that avoid local cognitive traps. Models with smaller parameter sizes like Qwen2.5VL-3B and InternVL3-2B exhibit significantly greater difficulty in extracting implicit cues compared to their larger counterparts. These models frequently commit errors in the early stages of CoT reasoning, compromising subsequent logical coherence. 
~\cref{fig:radargram} illustrates a typical visual comparison.

\begin{minipage}{\textwidth}
  \begin{minipage}[b]{0.65\textwidth}
        \centering 
        \scalebox{0.8}{
        \tabcolsep=0.09cm
        \begin{tabular}{c || c c c c c c}
        \hline
        \multirow{2}{*}{\bf \centering Method} & {\bf \raggedleft Street} & {\bf \raggedleft City} &{ \bf \raggedleft Region} & { \bf \raggedleft Country} & { \bf \raggedleft Continent} & { \bf \raggedleft CoT} \\ 
        
         & \bf $1$ km & \bf $25$ km & \bf $200$ km & \bf $750$ km & \bf $2500$ km & \bf quality\\ 
        
        \hline
        \hline

        ISNs & 1.76 & 11.23 & 16.94 & 23.08 & 26.4 & -\\

        GeoCLIP & 2.45 & 15.71 & 34.08 & 64.85 & 76.61 & -\\

        \hline
        
        InternVL2.5-4B & 0.05 & 2.74 & 5.09 & 12.08 & 18.96 & 31.22 \\
        
        InternVL2.5-8B  & 0.33 & 3.44 & 6.75 & 14.62 & 22.64 & 34.29 \\

        InternVL3-2B  & 0.19 & 0.75 & 1.56 & 3.82 & 6.18 & 23.41 \\

        InternVL3-8B  & 1.32 & 7.50 & 14.34 & 25.90 & 35.38 & 36.48\\
        
        Qwen2.5VL-3B  & 0.19 & 0.61 & 2.03 & 3.40 & 5.14 & 37.93\\
        
        Qwen2.5VL-7B & 0.33 & 4.34 & 6.84 & 9.39 & 10.90 & 50.36 \\  
        
        Qwen2.5VL-32B & 5.45 & 23.12 & 37.41 & 54.33 & 65.00 & 55.56 \\
        
        \bf Ours & \bf 6.14 & \bf 26.15 & \bf 44.67 & \bf 66.56 & \bf 83.16 & \bf 59.54\\
        
        \hline
        \end{tabular}
        }
        \captionof{table}{\small{Performance comparisons among traditional leading models, open-source MLLMs, and our GRE on GREval-Bench.}}
        \label{tab:bench}
  \end{minipage}
  \hfill
  \begin{minipage}[b]{0.33\textwidth}
    \centering
    \centerline{\includegraphics[width=\textwidth]{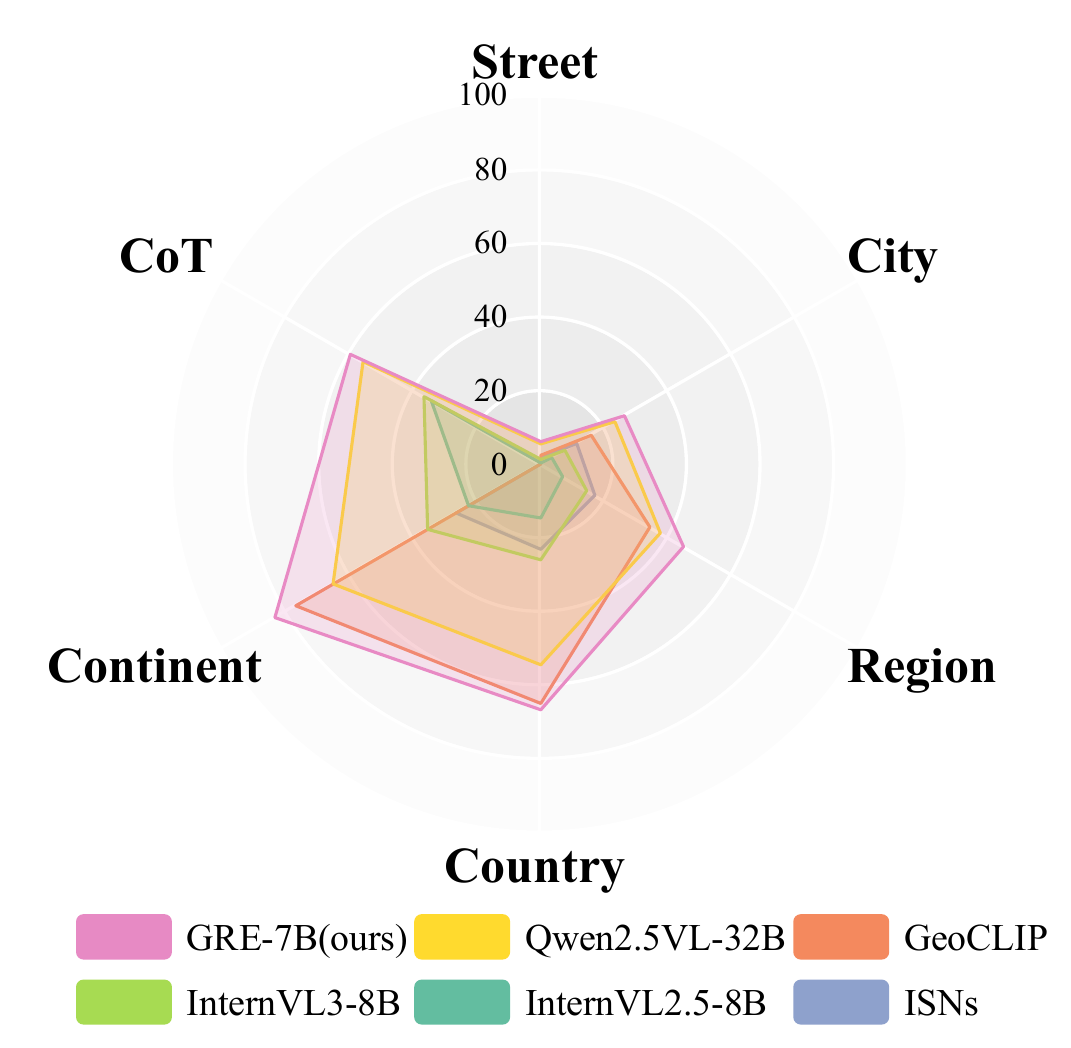}}
    \captionof{figure}{\small{Visual demonstration of the performance of models.}}
    \label{fig:radargram}
  \end{minipage}
  
\end{minipage}

\subsection{Ablation Study}
\vspace{-2mm}
 To evaluate the effectiveness of our training data and training strategies, we compare the model’s performance under four distinct training strategies: (1) applying Cold-start Initialization on our dataset, (2) further optimizing the GRE-CI with RL stage I, (3) further optimizing the GRE-CI with RL stage II, and (4) further optimizing the GRE-CI with RL stage I and stage II. As illustrated in ~\cref{tab:ablation}, the application of CI on our dataset significantly enhances the model’s performance on both the coarse-grained (e.g., country, continent) and fine-grained (e.g., city, street) localization performance. For (2) and (3) , (3) reach a comparable performance and (2) dropped at some levels of granularity, attributed to the misalignment between training and test task (reward) types in Stage I. Overall, (4) demonstrates superior performance to (3) due to its more robust reasoning capabilities. We also conduct additional ablation study on larger scale model and other open source model in Appendix~\ref{sec:additional_xperiment}, materials can be found in the repository.

\begin{table}[htp]
\centering 
\vspace{-0.1in}
\caption{\small{Ablation study on (a) Im2GPS3k~\cite{4587784} and (b) GWS15k~\cite{clark2023we} datasets.}} 
    \begin{subtable}{.46\linewidth}
    \raggedright
    \caption{\centering{\small{Results on the Im2GPS3k~\cite{4587784} dataset}}}
    \scalebox{0.7}
    {
    \tabcolsep=0.09cm
    \begin{tabular}{c || c c c c c}
    \hline
    \multirow{2}{*}{\bf \centering Method} & {\bf \raggedleft Street} & {\bf \raggedleft City} &{ \bf \raggedleft Region} & { \bf \raggedleft Country} & { \bf \raggedleft Continent} \\ 
    
     & \bf $1$ km & \bf $25$ km & \bf $200$ km & \bf $750$ km & \bf $2500$ km \\
    
    \hline
    \hline

    Qwen2.5-VL-7B & 3.20 & 16.62 & 28.03 & 42.14 & 52.99\\  
    
    CI & 7.77 & 29.30 & 44.78 & 62.43 & 78.81\\
    
    CI + I & 7.16 & 28.13 & 42.41 & 63.29 & 78.61\\
    
    CI + II & 10.96 & \underline{36.11} & \underline{52.17} & 67.26 & 83.32\\
    
    CI + I + II & \bf 11.33 & 35.28 & 51.72 & \bf 69.33 & \bf 85.67\\
    
    \hline
    \end{tabular}
    }
    \end{subtable}
\hskip0.8cm
    \begin{subtable}{.46\linewidth}
    \raggedleft
    \caption{\centering{\small{Results on the recent GWS15k~\cite{clark2023we} dataset}}}
    \scalebox{0.7}
    {
    \tabcolsep=0.09cm
    \begin{tabular}{c || c c c c c}
    \hline
    \multirow{2}{*}{\bf \centering Method} & {\bf \raggedleft Street} & {\bf \raggedleft City} &{ \bf \raggedleft Region} & { \bf \raggedleft Country} & { \bf \raggedleft Continent} \\ 
    
    & \bf $1$ km & \bf $25$ km & \bf $200$ km & \bf $750$ km & \bf $2500$ km \\
    
    \hline
    \hline
    \multirow{4}{0.1cm}
    
    Qwen2.5-VL-7B & 0.05 & 0.29 & 1.39 & 4.43 & 8.66\\  
    
    CI & 0.45 & 2.17 & 12.91 & 37.58 & 61.83\\
    
    CI + I & 0.35 & 2.03 & 12.82 & 37.88 & 62.16\\
    
    CI + II & 0.88 & 3.91 & 18.69 & \underline{55.61} & 78.03\\
    
    CI + I + II & \bf 0.91 & \bf 4.13 & \bf 18.86 & 54.82 & \bf 78.28\\
    
    \hline
    \end{tabular}
    }
    \end{subtable}
\label{tab:ablation}
\end{table}

\section{Conclusion}
\vspace{-3mm}
\label{conclusion}
\hyphenpenalty=300
\tolerance=100
In this paper, we introduce a comprehensive framework for visual geo-localization reasoning, built upon a formalization approach that unifies data construction, model training, and evaluation. Our framework is designed to address the limitations of the current methods, enabling model to reason in geo-localization task. The ability of extracting of multigranular visual cues from images and integrating with external world knowledge will also inspire us in other domains of VLMs. This framework has led to the creation of the GRE dataset, a rich resource featuring detailed step-by-step reasoning annotations designed to enhance model training and evaluation on geo-localization task. The GRE model, trained using this framework, demonstrates strong geo-localization reasoning capabilities and exhibits robust generalization across a diverse range of scenes, from implicit scenes to explicit scenes. To further support the evaluation of geo-localization, we introduce GREval-Bench, a comprehensive benchmark that rigorously assesses model performance across various geospatial scenario. Our extensive experiments validate the effectiveness of our approach, showing significant improvements over state-of-the-art open-source models. 

\small
\bibliographystyle{plainnat}
\bibliography{references}
\normalsize

\newpage
\appendix
\section{More Details of GRE30K}

\subsection{Example Illustrations of GRE30K}
\label{sec:gre30k_examples}
\begin{figure}[h!]
  \centering
  \includegraphics[width=\linewidth]{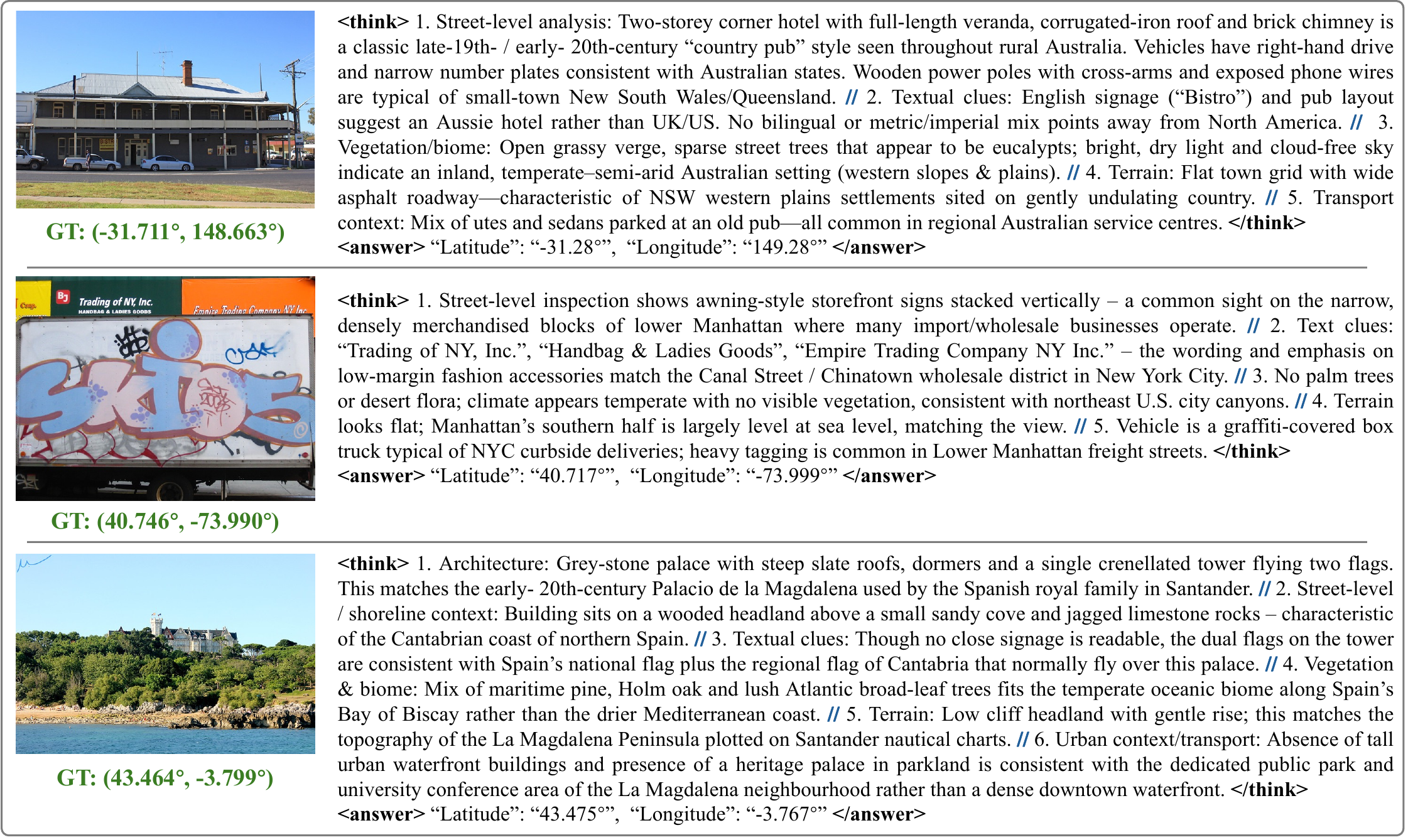}
  \caption{Three examples to show CoT data in GRE30K-CoT. }
  \label{fig:gre30k_cot_case}
\end{figure}

\begin{figure}[h!]
  \centering
  \includegraphics[width=\linewidth]{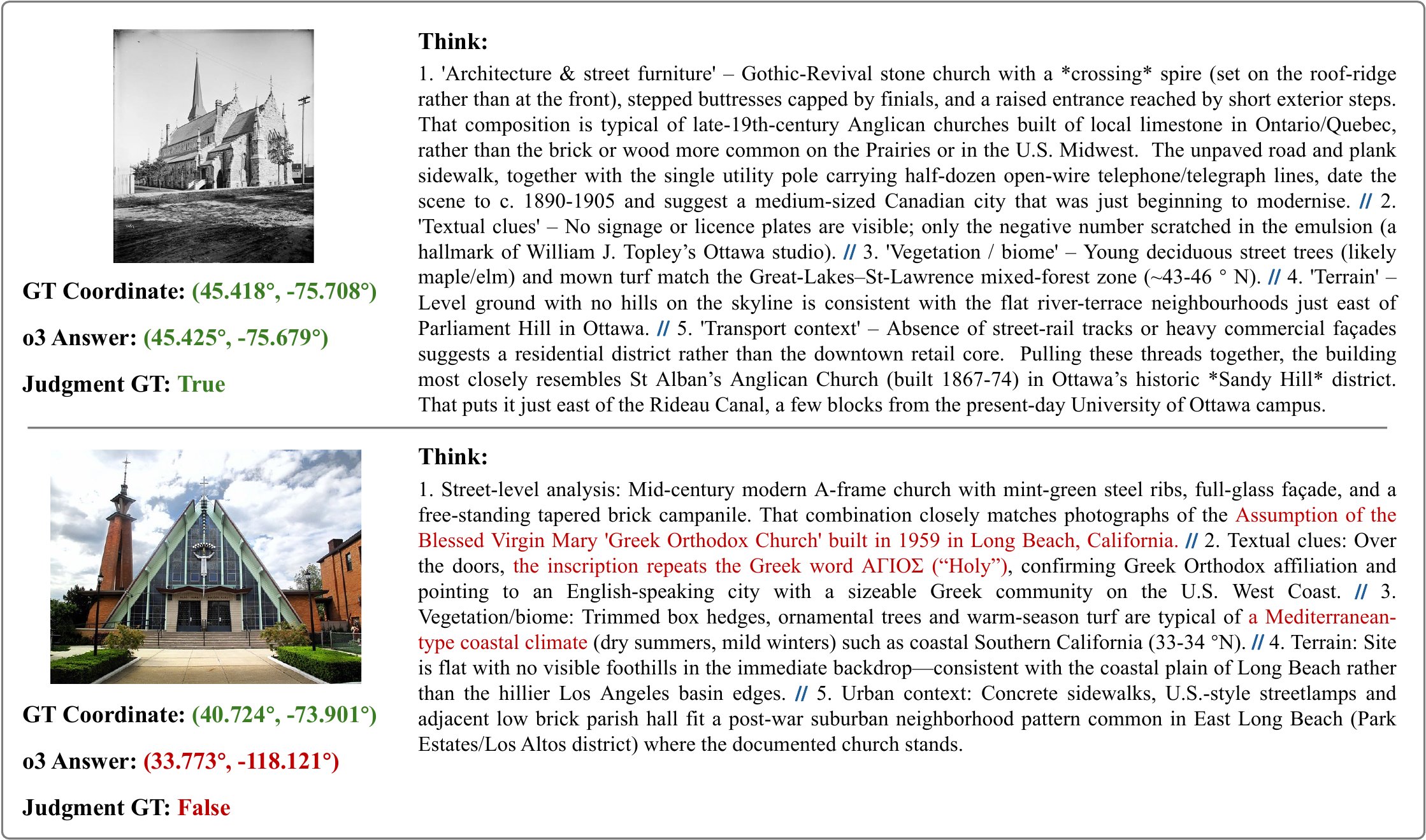}
  \caption{Two examples to show Judgment data in GRE30K-Judge. \textcolor[RGB]{190,38,15}{Red} option indicates the wrong reasoning steps.}
  \label{fig:gre30k_judge_case}
\end{figure}

We provide several examples to illustrate the reasoning data in our GRE30K dataset, including the
high-quality CoT data in ~\cref{fig:gre30k_cot_case} and the judgment data in ~\cref{fig:gre30k_judge_case}.

\subsection{Detailed prompt for GPT-o3}

\begin{figure}[h!]
  \centering
  \includegraphics[width=\linewidth]{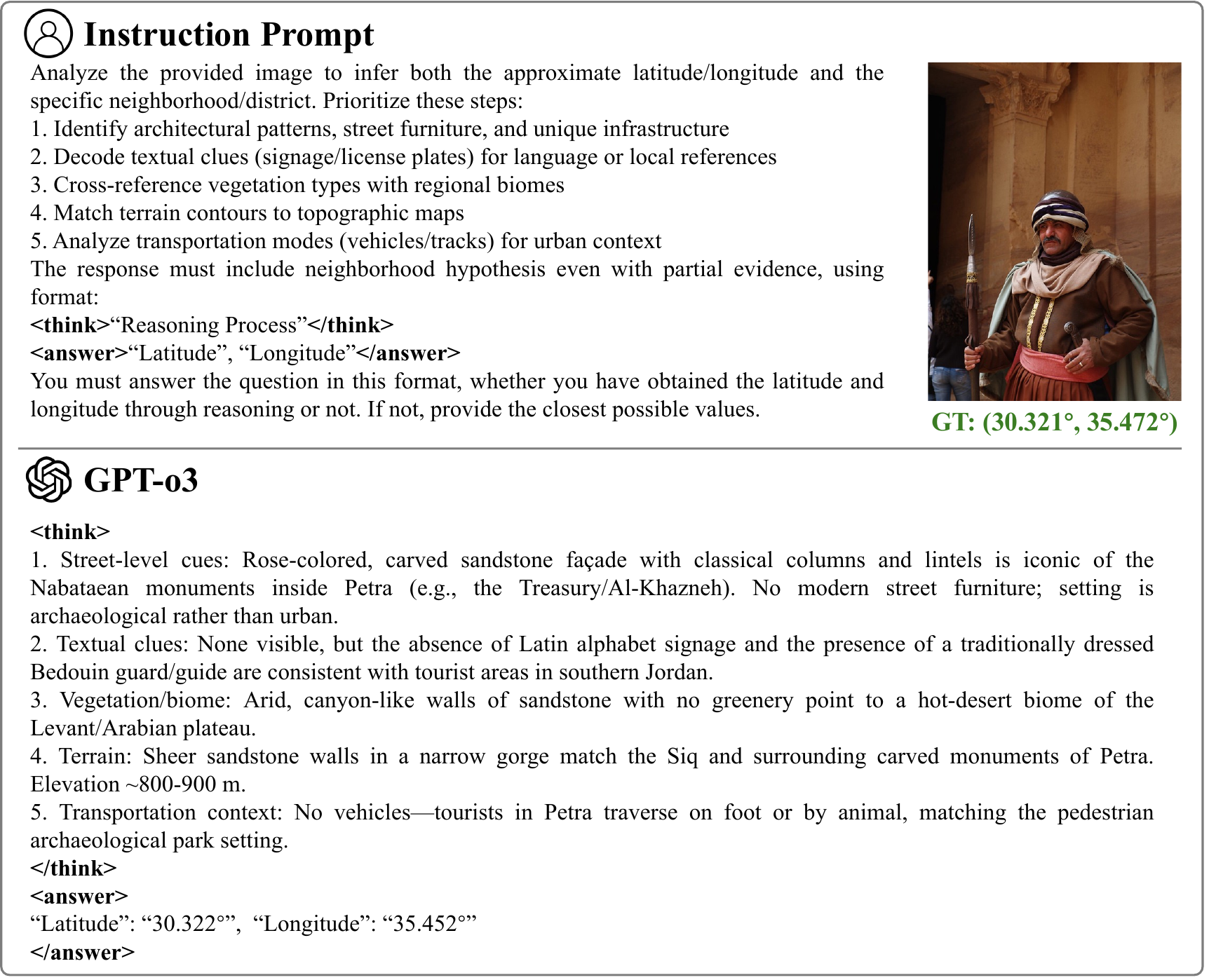}
  \caption{One example to illustrate the prompt for GPT-o3 to generate CoT data. The top block indicates the contexts including the image and instruction used to prompt o3, and the bottom block shows the response. }
  \label{fig:o3_prompt}
\end{figure}

Please refer to ~\cref{fig:o3_prompt} for more details.
\label{sec:o3_prompt}

\subsection{Review and Refinement Pipeline for GRE30K-CoT}
\label{sec:cot_details}
\textbf{Review and Refinement Pipeline for GRE30K-CoT.}
After the data generation process, we employ regular expression matching to filter out samples where the predicted coordinates deviate from the ground truth beyond a threshold $\theta$. Notably, these filtered samples are not discarded but rather incorporated into GRE30K-Judge.
To ensure the high quality of the generated samples, we apply manual verification after automated filtering. The process is conducted by three trained annotators with geographic-relevant professional backgrounds. The annotators examine and correct hallucinated image descriptions and inconsistent geographical reasoning in the CoT, ensuring that o3's output adheres to ``\texttt{<think> </think><answer> </answer>}'' format. Additionally, they maintain alignment between the reasoning process and the instruction structure.

\begin{figure}[h!]
  \centering
  \includegraphics[width=\linewidth]{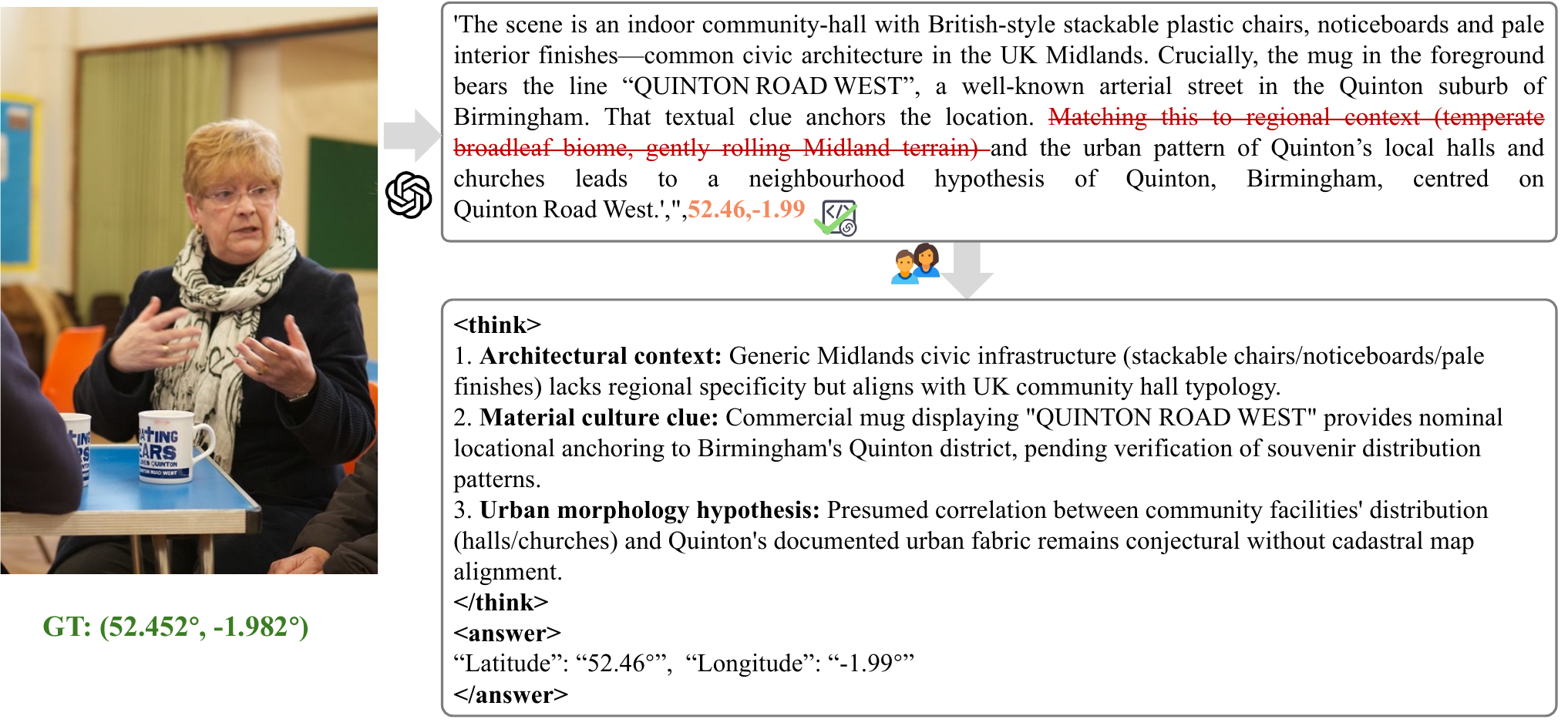}
  \caption{An illustrative example of Chain-of-Thought refinement and format normalization. The \textcolor[RGB]{190,38,15}{\sout{red strikethrough text}} denotes hallucinated content where the instructor model (o3) generated descriptions that are not actually present in the image.}
  \label{fig:manual}
\end{figure}

\textbf{Examples of Manual Filtering.}
As illustrated in ~\cref{fig:manual}, through a combination of regular expression matching and manual filtering, we enhance the quality of o3 generated Chain-of-Thought outputs, ultimately constructing a high-quality CoT dataset, GRE30K-CoT.

\section{More Details of GREval-Bench}
\subsection{Detail of GREval-Bench Construction and Evaluation Pipeline}
\label{sec:bench_details}
For image filtering, we construct a geographical reasoning corpus based on GRE30K-CoT, utilizing Named Entity Recognition (NER) to identify locations and architectural entities, and Semantic Role Labeling (SRL) to extract geographical reasoning patterns (e.g., ``spire style $\rightarrow$ European church''). The geographical indicators in the corpus are then categorized into explicit and implicit types. Explicit indicators encompass artificial landmarks, natural geographical features, and textual symbols, while implicit indicators include architectural styles, urban planning patterns, social characteristics, and environmental characteristics. Please refer to ~\cref{sec:subcategories_details} for detailed sub-categories. We employ CLIP~\cite{radford2021learning} to compute similarity scores between images and geography-relevant textual prompts from our geographical corpus (e.g., "base of Eiffel Tower", "Arabic text", "redwood forest"), retaining samples with high relevance scores. Subsequently, images with single facial regions occupying more than 50\% of the area are removed through face detection~\cite{schroff2015facenet}. The rule-filtered images then undergo manual verification, where annotators answer the question: ``Can the approximate geographical location (country/city level) be inferred solely from this image?'' Images are excluded if two or more out of three annotators respond negatively.

Inspired by previous CoT evaluation~\cite{chen2024m,yang2025r1,jiang2025mme}, we provide key steps annotation and reference GPS coordinate for all samples. We initially leverage o3 to generate the answer rationale. For the rationale, we provide both instructions and ground truth coordinates to o3. Subsequently, three geography domain annotators review and annotate key intermediate steps, utilizing o3's responses as reference. For cases where o3 fails to generate reasonable rationales, annotators develop geo-localization reasoning process independently. 

\subsection{Detailed Subcategories of Geographical Indicators}
\label{sec:subcategories_details}
In the image geolocation task, geolocation indicators refer to the visual elements in the image that can directly or indirectly infer the geographic location. ~\cref{tab:subcategories} shows the classification and specific examples of geolocation clues.

\begin{table}[h]
\centering
\caption{\small{Detailed subcategories of geographical indicators.}}
\scalebox{0.8}{
\tabcolsep=0.09cm
\begin{tabular}{c || c m{12cm}}
\hline
\bf Type & \bf Subcategory & \bf Scenario \\
\hline
\hline

Explicit  & \makecell[ct]{nature \\ landmark} & 
\begin{itemize}[leftmargin=*]
  \item \textbf{Global/National Landmarks}: Eiffel Tower (Paris); Statue of Liberty (New York); Great Wall (Beijing)
  
  \item \textbf{Regional Architecture}: Neuschwanstein Castle (Bavaria, Germany); Kiyomizu-dera Temple (Kyoto, Japan); Prague Astronomical Clock (Czech Republic)
  
  \item \textbf{Unique Structures}: Bridges (Golden Gate Bridge); Ferris Wheel (London Eye); Religious Buildings (Mosque Domes, Gothic Church Spires)
\end{itemize} 
\\
\hline
Explicit & \makecell[ct]{iconic \\ buildings}  & 
\begin{itemize}[leftmargin=*]
  \item \textbf{Global/National Landmarks}: Eiffel Tower (Paris); Statue of Liberty (New York); Great Wall (Beijing)
  
  \item \textbf{Regional Architecture}: Neuschwanstein Castle (Bavaria, Germany); Kiyomizu-dera Temple (Kyoto, Japan); Prague Astronomical Clock (Czech Republic)
  
  \item \textbf{Unique Structures}: Bridges (Golden Gate Bridge); Ferris Wheel (London Eye); Religious Buildings (Mosque Domes, Gothic Church Spires)
\end{itemize}
\\ 
\hline
Explicit & language & 
\begin{itemize}[leftmargin=*]
  \item \textbf{Language signs}: Language on road signs and store signs (Arabic $\rightarrow$ Middle Eastern; Cyrillic $\rightarrow$ Eastern European).
\end{itemize}
\\
\hline
Explicit & symbolic & 
\begin{itemize}[leftmargin=*]
  \item \textbf{Administrative signs}: License plates (German license plates "D")
  
  \item \textbf{Currency and flags}: Euro coins (European countries); Canadian maple leaf flag

\end{itemize}
\\
\hline
Implicit & \makecell[ct]{geographical \\ features}  & 
\begin{itemize}[leftmargin=*]
  \item \textbf{Unique landforms}: Uyuni Salt Flats (Bolivia); Grand Canyon (USA); Guilin Karst landforms
  
  \item \textbf{Vegetation types}: Cactus (desert areas); coconut trees (tropical coastal areas); birch trees (northern temperate zones)
  
  \item \textbf{Water features}: Victoria Falls (Africa); Dead Sea (high salinity water bodies)
\end{itemize}
\\
\hline
Implicit & \makecell[ct]{architectural \\ style}  & 
\begin{itemize}[leftmargin=*]
  \item \textbf{Architectural style}: Spanish colonial style (Mexico); neoclassicism (Washington, DC); earthen building (Fujian)
  
  \item \textbf{Street characteristics}: Narrow cobblestone roads (European ancient towns); grid layout (Manhattan, New York); tricycles (Southeast Asian cities)

\end{itemize}
\\
\hline
Implicit & \makecell[ct]{social \\ characteristics} & 
\begin{itemize}[leftmargin=*]
  \item \textbf{Clothing and customs}: Kimono (Japan); Scottish plaid skirt; Indian sari
  
  \item \textbf{Transportation}: Tunisian carriage; Venetian gondola; London red bus

\end{itemize}
\\
\hline
Implicit & climate & 
\begin{itemize}[leftmargin=*]
  \item \textbf{Seasons and Weather}: Aurora (high latitudes); monsoon rainforest (rainy season in Southeast Asia); sandstorms (deserts in the Middle East)
\end{itemize}
\\
\hline
\end{tabular}
}

\label{tab:subcategories}
\end{table}

\section{More Experiments}

\subsection{More Details on Training}
\label{sec:rl_training}
\begin{figure}[h!]
  \centering
  \includegraphics[width=\linewidth]{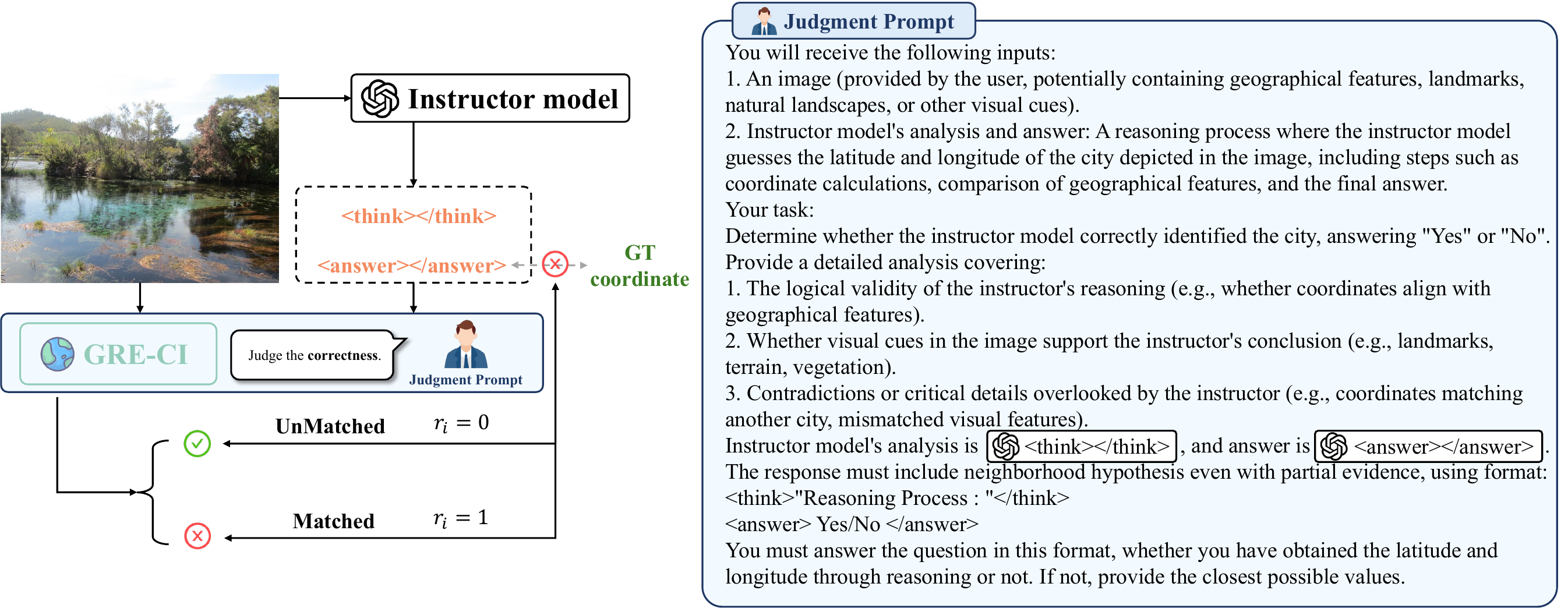}
  \caption{RL stage I training pipeline and Judgment Prompt.}
  \label{fig:rl1_prompt}
\end{figure}

\begin{figure}[h!]
  \centering
  \includegraphics[width=\linewidth]{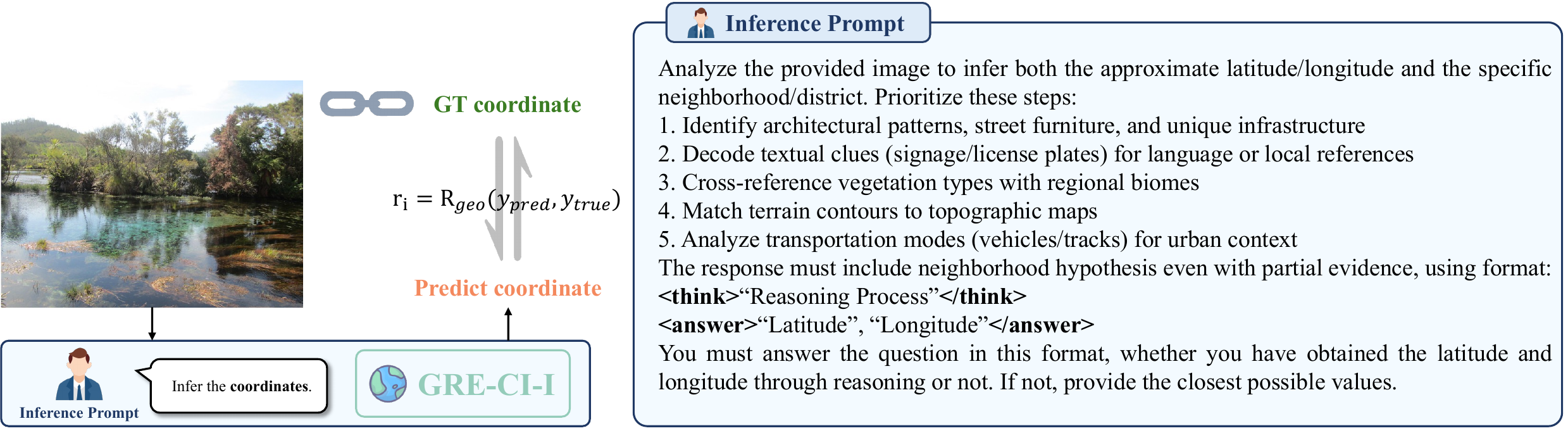}
  \caption{RL stage II training pipeline and Inference Prompt.}
  \label{fig:rl2_prompt}
\end{figure}

Please refer to ~\cref{fig:rl1_prompt} and ~\cref{fig:rl2_prompt} for more details. During the training process, the threshold is continuously updated. If the model can stably maintain enough rewards at the current granularity level, the threshold is further refined to a finer granularity level.

\subsection{Additional Main Results}
\label{sec:additional_xperiment}
We also conduct evaluations on the Google StreetView dataset(~\cref{tab:additional gsv}), where we observe similar performance trends. Additionally, we demonstrate the performance of our base model, Qwen-2.5VL series, on Im2GPS3k and GWS15k datasets(~\cref{tab:additional qwen2.5vl}). The results align with our conclusions from the main results, further validating the effectiveness of our proposed training strategy. We also have compared our model on the OSV-5M~\cite{10657636} in Table~\ref{tab:osv5m}, where our model emonstrates excellent performance.  As \textit{Around the World} demonstrates excellent performance, we conduct study on MP16 dataset, and materials can be found in the repository. 

\begin{table}[h]
\centering
\caption{\small{Results on the Google StreetView dataset.}}
\scalebox{0.8}{
\tabcolsep=0.09cm
\begin{tabular}{c || c c c c c}
\hline
\multirow{2}{*}{\bf \centering Method} & {\bf \raggedleft Street} & {\bf \raggedleft City} &{ \bf \raggedleft Region} & { \bf \raggedleft Country} & { \bf \raggedleft Continent}  \\ 

 & \bf $1$ km & \bf $25$ km & \bf $200$ km & \bf $750$ km & \bf $2500$ km \\

\hline
\hline

\multirow{4}{0.1cm}

\centering Qwen2.5VL-3B  & 4.47 & 46.92 & 68.22 & 78.26 & 83.89\\

\centering Qwen2.5VL-7B & 7.99 & 61.00 & 70.42 & 83.20 & 85.56\\ 

Qwen2.5VL-32B & 14.62 & 67.50 & 69.04 & 88.42 & 92.59 \\

CI & 15.53 & 64.25 & 74.46 & \underline{94.20} & \underline{96.14} \\

CI + I & 13.59 & 63.75 & 75.19 & 92.30 & 96.02 \\


\bf Ours & \bf 18.15 & \bf 71.01 & \bf 75.36 & 91.30 & 92.75 \\

\hline
\end{tabular}
}

\label{tab:additional gsv}
\end{table}

\begin{table}
\centering 
\vspace{-0.1in}
\caption{\small{We test the Qwen2.5VL series on (a) Im2GPS3k~\cite{4587784} and (b) GWS15k~\cite{clark2023we} datasets for reference here.}} 
    \begin{subtable}{.46\linewidth}
    \raggedright
    \caption{\centering{\small{Results on the Im2GPS3k~\cite{4587784} dataset}}}
    \scalebox{0.7}
    {
    \tabcolsep=0.09cm
    \begin{tabular}{c || c c c c c}
    \hline
    \multirow{2}{*}{\bf \centering Method} & {\bf \raggedleft Street} & {\bf \raggedleft City} &{ \bf \raggedleft Region} & { \bf \raggedleft Country} & { \bf \raggedleft Continent} \\ 
    
     & \bf $1$ km & \bf $25$ km & \bf $200$ km & \bf $750$ km & \bf $2500$ km \\
    
    \hline
    \hline

     Qwen2.5VL-3B & 0.33 & 1.20 & 3.57 & 5.37 & 7.31\\  
    
     Qwen2.5VL-7B & 3.20 & 16.62 & 28.03 & 42.14 & 52.99 \\
     
     Qwen2.5VL-32B & \bf 6.47 & \bf 25.12 & \bf 40.96 & \bf 59.87 & \bf 75.32\\
    
    \hline
    \end{tabular}
    }
    \end{subtable}
\hskip0.8cm
    \begin{subtable}{.46\linewidth}
    \raggedleft
    \caption{\centering{\small{Results on the recent GWS15k~\cite{clark2023we} dataset}}}
    \scalebox{0.7}
    {
    \tabcolsep=0.09cm
    \begin{tabular}{c || c c c c c}
    \hline
    \multirow{2}{*}{\bf \centering Method} & {\bf \raggedleft Street} & {\bf \raggedleft City} &{ \bf \raggedleft Region} & { \bf \raggedleft Country} & { \bf \raggedleft Continent} \\ 
    
     & \bf $1$ km & \bf $25$ km & \bf $200$ km & \bf $750$ km & \bf $2500$ km \\
    
    \hline
    \hline
    \multirow{4}{0.1cm}
    
     Qwen2.5VL-3B & 0.02 & 0.17 & 0.41 & 2.14 & 6.70 \\  
    
     Qwen2.5VL-7B & 0.05 & 0.29 & 1.39 & 4.43 & 8.66 \\
     
     Qwen2.5VL-32B & \bf 0.06 & \bf 0.36 & \bf 7.53 & \bf 28.46 & \bf 52.39\\
    
    
    
    \hline
    \end{tabular}
    }
    \end{subtable}

\label{tab:additional qwen2.5vl}
\end{table}

\begin{table}
        \centering 
         \caption{\centering{\small{Results on the recent OSV-5M~\cite{10657636} dataset}}}
        \scalebox{0.8}{
        \tabcolsep=0.09cm
        \begin{tabular}{c || c c c c c c}
        \hline
        \multirow{2}{*}{\bf \centering Method} & {\bf \raggedleft Street} & {\bf \raggedleft City} &{ \bf \raggedleft Region} & { \bf \raggedleft Country} & { \bf \raggedleft Continent} & { \bf \raggedleft dist} \\ 
        
         & \bf $1$ km & \bf $25$ km & \bf $200$ km & \bf $750$ km & \bf $2500$ km & \bf 	Average Distance\\ 
        
        \hline
        \hline

        Qwen2.5VL-7B & 1.0 & 1.9 & 4.8 & 19.0 & 43.1 & 4942 \\
        Molmo-D-7B & 0.7 & 1.1 & 1.3 & 7.2 & 32.1 & 6172 \\
        LLaVA-V1.5-7B & 0.1 & 0.2 & 0.7 & 5.0 & 21.9 & 6895 \\
        SeekWorld & 1.0 & 1.3 & 7.0 & 27.6 & 51.3 & 4326 \\
        SC Retrieval & - & \underline{19.9} & \underline{45.8} & 73.4 & - & 1386 \\
        RFM $\mathcal{S}_2$ & - & 5.4 & 44.2 & \underline{76.2} & - & \underline{1069} \\
        \bf Ours & \textbf{5.7} & 9.7 & 35.57 & 72.53 & \textbf{91.11} & 1192 \\
        
        \hline
        \end{tabular}
        }
        \label{tab:osv5m}
\end{table}

\subsection{Additional Ablation Study}

We also conduct additional ablation study on Qwen2.5VL-32B and LLaVA-v1.5-7B~\cite{GRPO4llava}(Table ~\ref{tab:moreablation}), where we observe similar
performance trends. The results demonstrate the efficacy and broad applicability of the proposed method.

\begin{table}[htp]
\centering 
\vspace{-0.1in}
\caption{\small{More ablation study on (a) Im2GPS3k~\cite{4587784} and (b) GWS15k~\cite{clark2023we} datasets.}} 
    \begin{subtable}{.46\linewidth}
    \raggedright
    \caption{\centering{\small{Results on the Im2GPS3k~\cite{4587784} dataset}}}
    \scalebox{0.7}
    {
    \tabcolsep=0.09cm
    \begin{tabular}{c || c c c c c}
    \hline
    \multirow{2}{*}{\bf \centering Method} & {\bf \raggedleft Street} & {\bf \raggedleft City} &{ \bf \raggedleft Region} & { \bf \raggedleft Country} & { \bf \raggedleft Continent} \\ 
    
     & \bf $1$ km & \bf $25$ km & \bf $200$ km & \bf $750$ km & \bf $2500$ km \\
    
    \hline
    \hline

    LLaVA-v1.5-7B & 1.7 & 7.5 & 11.3 & 20.8 & 44.6 \\
    

    CI(LLaVA) & 4.2 & 10.2 & 24.9 & 42.9 & 58.9\\

    CI + II(LLaVA) & 6.1 & 14.6 & 31.3 & 47.6 & 63.1\\

    Qwen2.5-VL-32B & 6.5 & 25.1 & 41.0 & 59.9 & 75.3\\

    CI(Qwen) & 8.1 & 31.4 & 46.5 & 69.7 & 81.1\\
    
    CI + I(Qwen) & 7.6 & 30.1 & 40.2 & 71.2 & 82.2\\
    
    CI + I + II(Qwen) & \bf 12.3 & \bf 36.6 & \bf 59.3 & \bf 78.3 & \bf 88.6\\
    
    \hline
    \end{tabular}
    }
    \end{subtable}
\hskip0.8cm
    \begin{subtable}{.46\linewidth}
    \raggedleft
    \caption{\centering{\small{Results on the recent GWS15k~\cite{clark2023we} dataset}}}
    \scalebox{0.7}
    {
    \tabcolsep=0.09cm
    \begin{tabular}{c || c c c c c}
    \hline
    \multirow{2}{*}{\bf \centering Method} & {\bf \raggedleft Street} & {\bf \raggedleft City} &{ \bf \raggedleft Region} & { \bf \raggedleft Country} & { \bf \raggedleft Continent} \\ 
    
    & \bf $1$ km & \bf $25$ km & \bf $200$ km & \bf $750$ km & \bf $2500$ km \\
    
    \hline
    \hline
    \multirow{4}{0.1cm}
    
    Qwen2.5-VL-32B & 0.06 & 0.36 & 7.5 & 28.5 & 52.4\\ 
    
    CI & 0.51 & 2.8 & 15.1 & 43.4 & 68.1\\
    
    CI + I & 0.42 & 2.3 & 13.9 & 43.6 & 68.1\\
    
    CI + I + II & \bf 0.97 & \bf 4.9 & \bf 20.1 & \bf 57.0 & \bf 81.3\\
    
    \hline
    \end{tabular}
    }
    \end{subtable}
\label{tab:moreablation}
\end{table}

\subsection{Qualitative Results}

In the supplementary materials, we provide additional visual examples illustrating the reasoning performance on the image geo-localization task. These examples demonstrate GRE's capability to generate remarkable chains of thought for accurate coordinate prediction in challenging scenarios.

\section{Limitations and Future Work}

\subsection{Limitations}
\label{sec:limitations}
The primary limitations of GRE include (1) substantial computational resource requirements, specifically utilizing 8 NVIDIA H20 GPUs for model training, and (2) the associated API costs for dataset generation. GeoCLIP requires 155.63 GFLOPs per inference. In comparison, our model requires 262.27 GFLOPs for the visual encoder and 24,117.47 GFLOPs for the language model, which corresponds to 13.0506 GFLOPs per token. All FLOPs are measured using the THOP package.

\subsection{Future Work}
\label{sec:future_work}
Leveraging geo-localization reasoning capabilities, we can implement geographic information privacy identification and protection mechanisms. Furthermore, this approach can be extended through agent-based architectures that integrate reasoning capacities with tool invocation functionalities.

\subsection{Broader Impacts}
\label{sec:broader_impacts}
The reasoning capacity improvement in geo-localization facilitates the extraction of multi-granularity geographic indicators from imagery, offering dual benefits for geospatial data mining applications and location privacy preservation frameworks.

\section{More Qualitative Results}

\begin{figure}[h!]
  \centering
  \includegraphics[width=\linewidth]{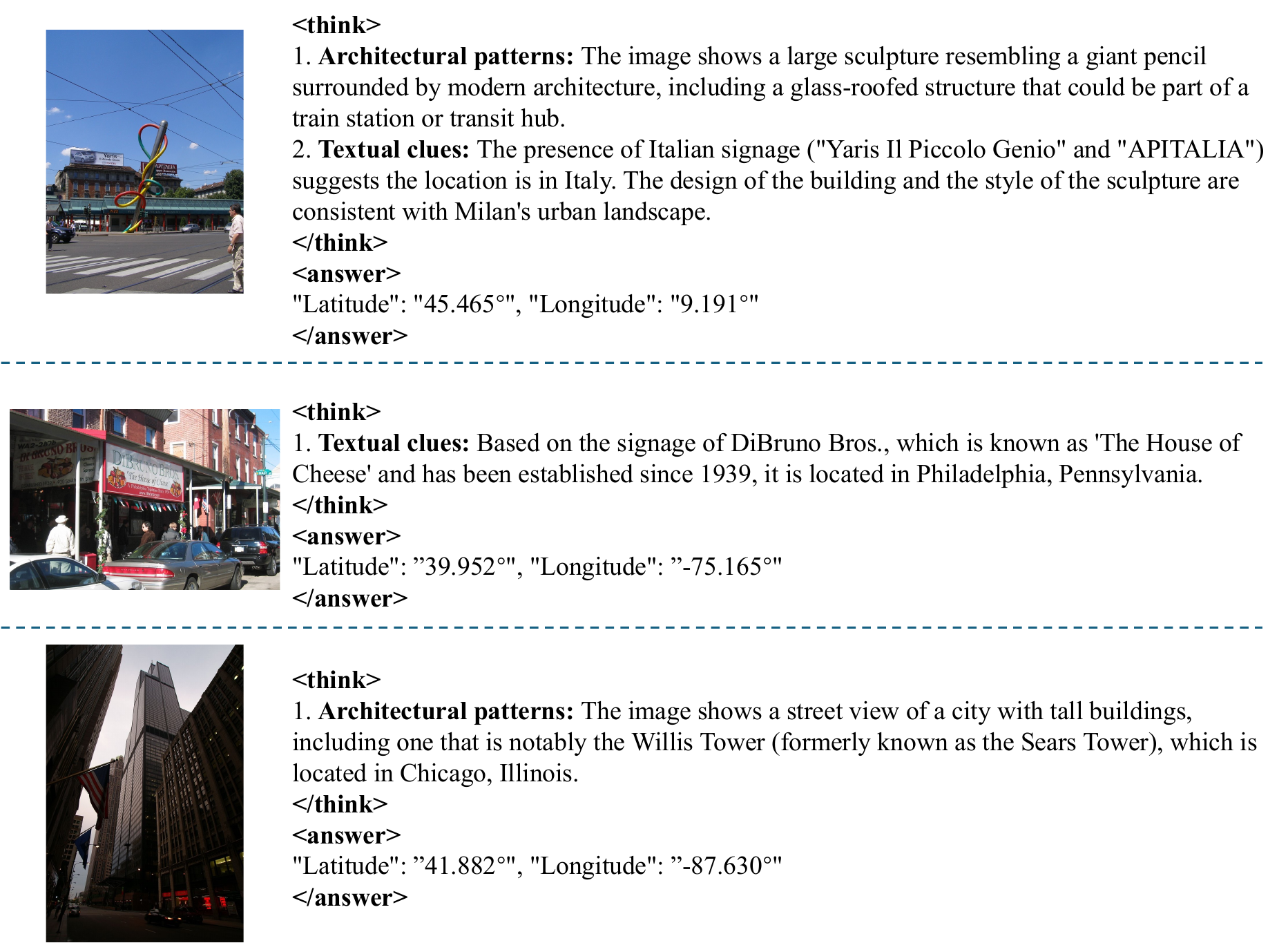}
  \caption{Visual examples of GRE.}
  \label{fig:more_cases}
\end{figure}

\begin{figure}[h!]
  \centering
  \includegraphics[width=\linewidth]{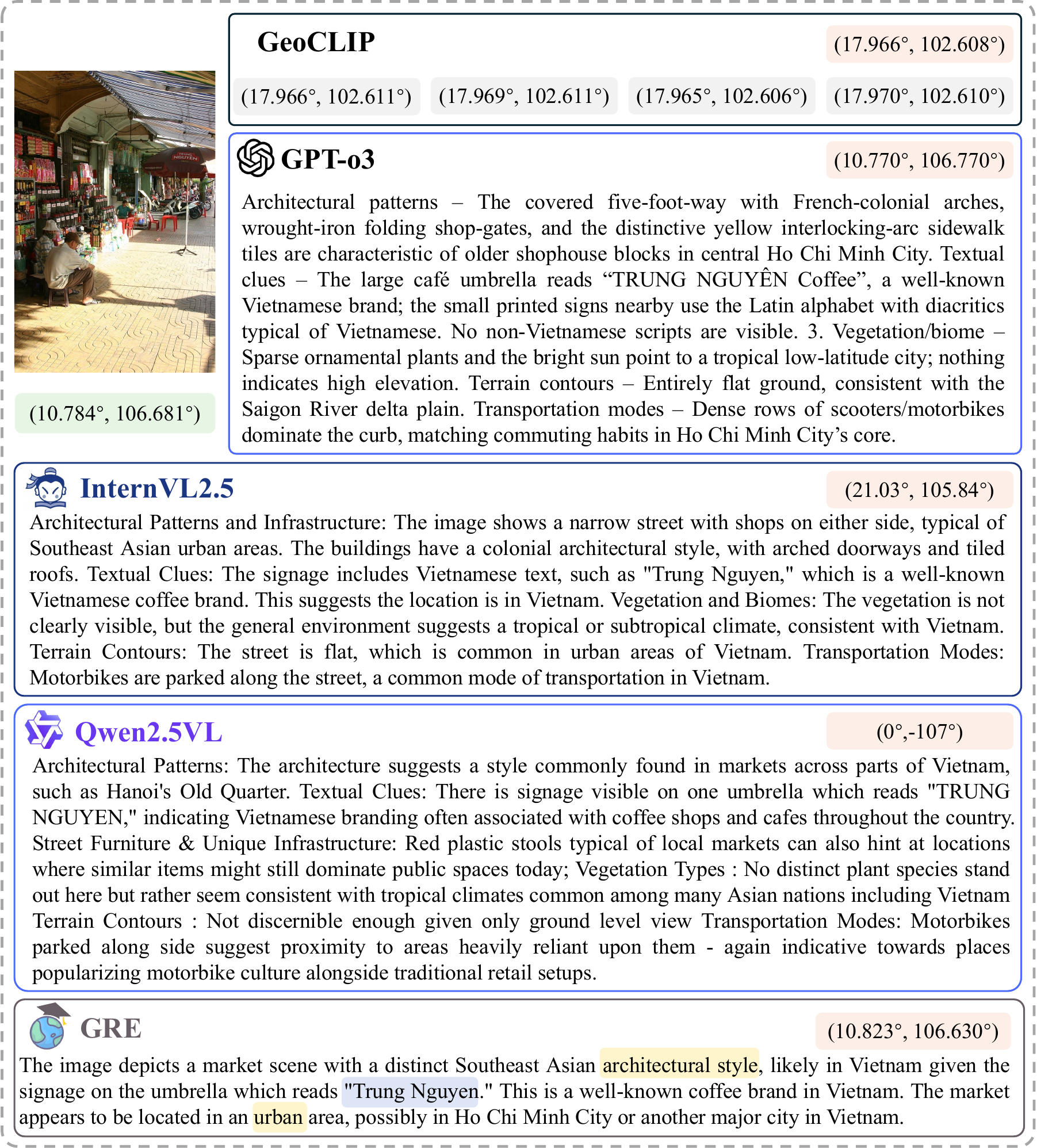}
  \caption{Qualitative comparisons with previous alignment-based methods and existing MLLMs with reasoning capabilities. \colorbox[RGB]{233,244,229}{$(Lat,Lon)$} denotes the ground truth coordinates, \colorbox[RGB]{251,239,232}{$(Lat,Lon)$} denotes the models' predicted answer, \colorbox[RGB]{223,229,245}{Indicator} denotes the explicit indicator and \colorbox[RGB]{253,245,208}{Indicator} denotes the implicit indicator. Notably, GeoCLIP generate five \colorbox[RGB]{242,242,242}{candidates coordinates} and select the candidate with the maximum probability score as the answer.}
  \label{fig:comparision_1}
\end{figure}

\begin{figure}[h!]
  \centering
  \includegraphics[width=\linewidth]{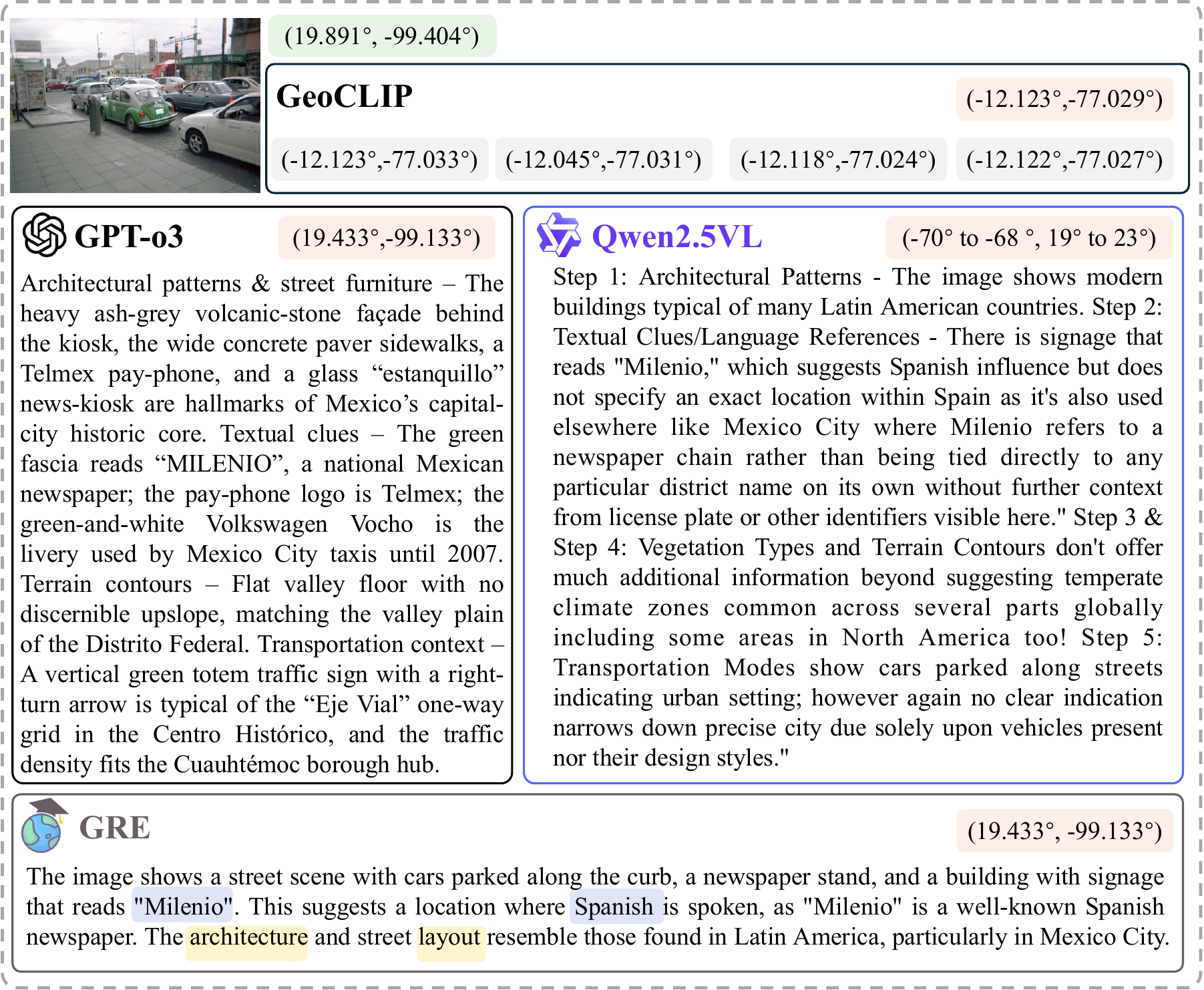}
  \caption{Qualitative comparisons.}
  \label{fig:comparision_2}
\end{figure}

We present additional visual examples to highlight the geographic reasoning performance. ~\cref{fig:more_cases} displays more
visual cases involving diverse locations. GRE is capable to generate explainable predictions with robust capabilities in these challenging scenarios. Furthermore, ~\cref{fig:comparision_1} and ~\cref{fig:comparision_2} provides comparisons with previous alignment-based methods and existing MLLMs with reasoning capabilities.
Our approach exhibits superior image geo-localization results
with implicit geographic indicators.

\end{document}